\documentclass{article}
\usepackage{iclr2027_conference,times}
\usepackage{amsmath,amssymb,amsthm,mathtools}
\newcommand{\R}{\mathbb{R}}

\newcommand{\E}{\mathbb{E}}
\newcommand{\1}{\mathbf{1}}

\newcommand{\cA}{\mathcal{A}}
\newcommand{\cL}{\mathcal{L}}

\newcommand{\cZ}{\mathcal{Z}}

\newcommand{\ip}[2]{\left\langle #1,#2\right\rangle}
\newcommand{\norm}[1]{\left\lVert #1\right\rVert}
\newcommand{\abs}[1]{\left\lvert #1\right\rvert}

\newcommand{\push}{{}_{\#}}
\DeclareMathOperator*{\argmin}{arg\,min}

\DeclareMathOperator{\atanTwo}{atan2}

\usepackage{booktabs,multirow,makecell}
\usepackage{algorithm}
\usepackage[noend]{algpseudocode}
\usepackage{graphicx}
\usepackage{subcaption}
\usepackage{tikz}
\usetikzlibrary{arrows.meta,calc,fit,positioning,shapes.geometric}
\usepackage{xcolor}
\usepackage{microtype}
\usepackage{placeins}
\usepackage{enumitem}
\usepackage{hyperref}
\hypersetup{hidelinks,pdftitle={Horizontal Flow Matching for Single-Cell Spatiotemporal Dynamics}}
\usepackage{url}

\newtheorem{theorem}{Theorem}
\newtheorem{proposition}[theorem]{Proposition}
\newtheorem{lemma}[theorem]{Lemma}
\newtheorem{corollary}[theorem]{Corollary}

\newcommand{\method}{\mbox{\textsc{Horizon}}}
\newcommand{\qWFR}{\overline{\mathrm{WFR}}}
\newcommand{\WFR}{\mathrm{WFR}}
\newcommand{\KL}{\mathrm{KL}}
\newcommand{\SE}{\mathrm{SE}}
\newcommand{\dd}{\mathrm{d}}
\newcommand{\cmark}{\checkmark}
\newcommand{\best}[1]{\textbf{#1}}
\newcommand{\second}[1]{\underline{#1}}
\newcommand{\rev}[1]{#1}

\definecolor{gaugeblue}{HTML}{31688E}
\definecolor{gaugeteal}{HTML}{35B779}
\definecolor{gaugeorange}{HTML}{F28E2B}
\definecolor{gaugered}{HTML}{D1495B}
\definecolor{gaugegray}{HTML}{6B7280}

\title{Horizontal Flow Matching\\
for Single-Cell Spatiotemporal Dynamics}
\author{Zihan Wang\textsuperscript{1}\thanks{Equal contribution.}\,,
  Jianzhe Li\textsuperscript{2}\footnotemark[1]\,,
  Zhenyi Zhang\textsuperscript{3}\thanks{Corresponding authors: \texttt{zhenyizhang@stu.pku.edu.cn}, \texttt{pjzhou@pku.edu.cn}.}\ \ \&
  Peijie Zhou\textsuperscript{1,2,4,5}\footnotemark[2] \\
  \\
  \textsuperscript{1}Center for Quantitative Biology, Peking University \\
  \textsuperscript{2}Center for Machine Learning Research, Peking University \\
  \textsuperscript{3}LMAM and School of Mathematical Sciences, Peking University \\
  \textsuperscript{4}National Engineering Laboratory for Big Data Analysis and Applications, Beijing \\
  \textsuperscript{5}AI for Science Institute, Beijing
}

\iclrfinalcopy

\begin{document}
\maketitle
\raggedbottom
\setlength{\textfloatsep}{9pt plus 2pt minus 2pt}
\setlength{\floatsep}{7pt plus 2pt minus 2pt}
\setlength{\intextsep}{9pt plus 2pt minus 2pt}
\setlength{\abovecaptionskip}{5pt}
\makeatletter
\setlength{\@fptop}{0pt}
\makeatother

\begin{abstract}
Learning the spatiotemporal dynamics of cells from spatial transcriptomics
requires connecting snapshots of different cells acquired in different
coordinate frames.
A rotation or translation of a section can appear as cell movement and
distort the inferred correspondence.  We introduce horizontal flow matching,
which learns changes in tissue organization, molecular state, and cell
abundance up to shared rigid motion.  The dynamics follow a
continuity--reaction equation whose action assigns no cost to shared rigid
motion, and the resulting horizontal projector removes this component from
spatial prediction errors during learning and from predicted velocities
during generation.  An equivalent endpoint problem jointly estimates
correspondence and relative pose in Wasserstein--Fisher--Rao geometry and
yields analytic paths whose velocity and growth serve as regression targets.
Regressing these targets recovers the horizontal dynamics under ideal
population conditions.  Our model,
\method{}, uses an equivariant transformer to predict velocity and growth
from the current tissue, so that a tissue can be generated continuously from
a single initial snapshot.  On a simulated tissue observed in perturbed
coordinate frames, alignment improves reconstruction and projection reduces
rigid drift.  On four spatial transcriptomics time courses, \method{} reconstructs
tissue geometry, gene expression, and cell abundance more accurately than
existing spatial and unbalanced trajectory methods, and interpolates held-out
developmental stages.
\end{abstract}

\section{Introduction}

Development and regeneration change the composition of a tissue, the
molecular state of its cells, and their spatial arrangement at the same
time.  A generative model of these coupled changes would describe how a
tissue changes continuously from one observed stage to the next, a step
toward virtual tissues~\citep{bunne2024virtualcell,cao2026virtualembryo}.
Spatial transcriptomics records cell positions together with gene
expression, but the measurement destroys the
sample~\citep{stahl2016visualization,chen2022stereoseq,moses2022museum}.
A time course therefore consists of snapshots of different cells, and the
dynamics connecting them are not
observed~\citep{weinreb2018fundamental,schiebinger2019optimal}. In spatial transcriptomics, tissue sections from different stages are also recorded in independent
coordinate frames, introducing a geometric ambiguity.  A generative model must therefore distinguish changes in
the tissue itself from changes in the frame.

Dynamic optimal transport (OT) connects  distributions by the path of
least kinetic action~\citep{benamou2000computational,villani2009optimal},
and flow matching learns such paths by regressing velocity targets along
conditional interpolants~\citep{lipman2023flow,liu2023rectified}.
Trajectory inference applies these ideas to single-cell time
courses~\citep{tong2020trajectorynet,huguet2022mioflow}.  When abundance
changes, unbalanced OT also allows mass to be created and
destroyed~\citep{chizat2018unbalanced,liero2018oet}, and WFR-FM regresses
velocity and growth jointly along closed-form Wasserstein--Fisher--Rao (WFR)
paths~\citep{peng2026wfrfm}. However, these formulations describe dynamics in the recorded coordinates and do not
separate tissue deformation from a shared change of frame.

This ambiguity is specific to spatial measurements, since sections collected
at different times are treated separately and can differ by
rotations and translations unrelated to biological
change~\citep{zeira2022paste,clifton2023stalign}. Unaligned frames
distort cell correspondence during matching,
make rigid frame motion appear as cellular movement, and induce rigid drift
when the learned velocity is integrated. Pre-aligning the sections addresses
the first two issues but does not constrain rigid components that may reappear
during generation. stVCR jointly learns rigid alignment and neural ODE dynamics by
simulation~\citep{peng2026stvcr,chen2018neuralode}, while \mbox{STORIES}
avoids a shared frame by comparing within-section distances but does not learn
spatial velocity~\citep{huizing2026stories}. Quotient-space diffusion removes
rigid motion from updates of fixed atomic structures~\citep{xu2026quotient},
and \rev{rigid-motion-invariant} dynamic OT removes it from transport between
fixed-mass measures~\citep{toukam2026sod}. Tissue time courses additionally
involve different cells, changing mass, and coupled spatial and molecular
dynamics. This setting therefore requires a notion of motion relative to the
tissue itself that is compatible with unbalanced transport and used
consistently in the training paths, regression loss and integrated velocity.

\begin{figure*}[t]
\centering
\input{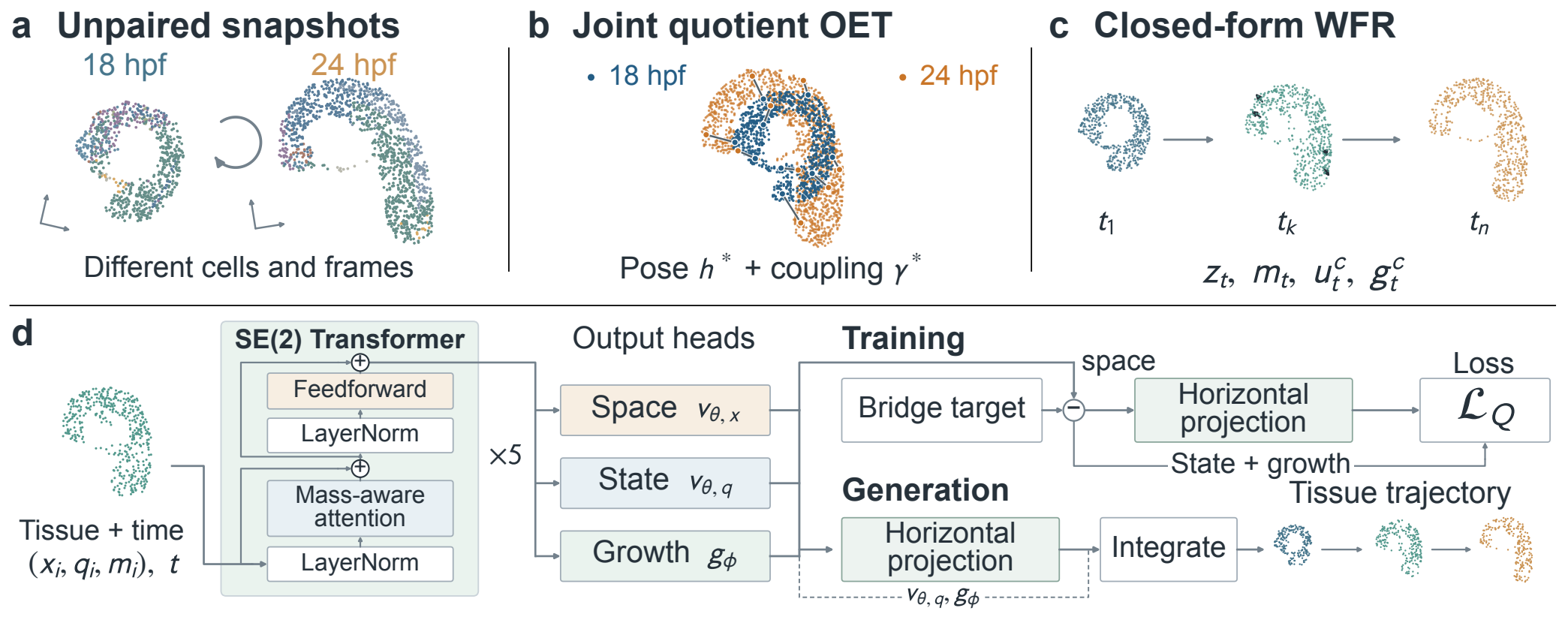}
\caption{\textbf{Horizontal flow matching with \method{}.}
\textbf{(a)}~Snapshots contain
different cells in unrelated frames.  \textbf{(b)}~Quotient optimal entropy
transport (qOET) jointly estimates a rigid pose and an unbalanced
correspondence between cells.  \textbf{(c)}~Closed-form WFR bridges give
intermediate states, masses, velocities, and growth rates.
\textbf{(d)}~An equivariant transformer learns velocity and growth from these
states, and the horizontal projector is applied both in the spatial loss and
to the velocity used for integration.  Panels (a--c) use the 18--24 hpf
zebrafish interval.}
\label{fig:overview}
\end{figure*}

We introduce \emph{horizontal flow matching} to learn tissue dynamics up to
shared rigid motion. Starting from WFR continuity--reaction dynamics, we
assign no cost to a moving rigid frame, which yields a horizontal projector
that removes the best-fitting translation and rotation from the spatial
velocity. The same projector is used in both training and generation, so the
model captures changes in tissue organization while factoring out
unidentifiable frame motion. We show that the resulting dynamic problem is
equivalent to jointly estimating endpoint correspondence and relative pose,
yielding closed-form WFR paths whose velocity and growth provide regression
targets without ODE solves, while projected regression recovers the horizontal
dynamics in the population limit. Our model, \method{}, combines these
targets with an equivariant transformer to generate continuous tissue
trajectories from the initial snapshot alone
(Figure~\ref{fig:overview}). In our experiments, endpoint alignment improves
reconstructed tissue shape, while horizontal projection suppresses rigid
drift during generation. Our main contributions are threefold.
\begin{itemize}[leftmargin=*,itemsep=0pt,topsep=1pt]
  \item Horizontal flow matching defines tissue
  dynamics up to shared rigid motion, using a single projector consistently in
  spatial regression and generation while learning molecular change and cellular \rev{population} growth.

  \item The quotient WFR dynamic problem
  reduces to endpoint alignment, with qOET jointly estimating relative pose and
  correspondence, and projected regression recovers the resulting horizontal
  dynamics at the population optimum.

  \item \method{} combines joint alignment,
  closed-form WFR bridge targets, and equivariant prediction, and we evaluate it on
  a benchmark with hidden frame perturbations and on four biological series
  to separate the effects of alignment and projection.
\end{itemize}

\section{Related work}
\label{sec:related-work}

\paragraph{Transport and flow matching for cell dynamics.}
Dynamic optimal transport describes the path between two distributions
through a continuity equation and a kinetic
action~\citep{benamou2000computational,villani2009optimal}, and unbalanced
transport extends the description to populations whose mass
changes~\citep{chizat2018interpolating,chizat2018unbalanced,liero2016hk,kondratyev2016radon}.
Flow matching learns a velocity field by regressing targets along
conditional
paths~\citep{lipman2023flow,liu2023rectified,albergo2023stochastic,tong2023minibatch}.
In single-cell biology, transport connects snapshots of different
cells~\citep{schiebinger2019optimal,lavenant2024trajectory,klein2025moscot},
dynamical models learn the connecting
fields~\citep{tong2020trajectorynet,huguet2022mioflow,neklyudov2023action,neklyudov2024wlf},
and growth enters through unbalanced, interaction-aware, or branching
formulations~\citep{fischer2019inferring,sha2024tigon,zhang2025deepruot,zhang2025cytobridge,pariset2023unbalanced,tong2024simulationfree}\rev{, including
discrete branching~\citep{ying2026beyond}}.
WFR-FM learns movement and growth from closed-form WFR
geodesics~\citep{peng2026wfrfm,liero2018oet}, and related methods fit
velocity and growth jointly or adapt paths to data
geometry and lineage constraints~\citep{wang2025joint,kapusniak2024metric,wang2026phylofm}.  Our formulation
penalizes only spatial motion relative to the tissue.

\paragraph{Dynamics and registration in spatial transcriptomics.}
stVCR pairs an initial OT-based alignment with rigid transformations that
are optimized jointly with neural ODE dynamics~\citep{peng2026stvcr}.
\mbox{STORIES} learns a cell fate potential with fused Gromov--Wasserstein
(FGW) transport~\citep{memoli2011gromov,vayer2020fused}, in which molecular
discrepancies form the linear cost and a quadratic term compares
within-section spatial distances~\citep{huizing2026stories}.  These
distances do not change under rigid motion of a section, so \mbox{STORIES} needs no
common frame, but it also learns no spatial velocity.  NicheFlow models
transitions between spatial microenvironments~\citep{sakalyan2025nicheflow}\rev{,
and multi-marginal flow matching with learnt interpolants has been applied
to spatial transcriptomics~\citep{kviman2026multimarginal}}.
FlowCloud decodes point clouds from a neural ODE latent space trained with
Chamfer and Sinkhorn losses and needs rigid pre-alignment when sections
are rotated by more than \(30^\circ\)~\citep{liu2026flowcloud}.  Registration
methods such as PASTE, PASTE2, GPSA, and STalign align observed
sections~\citep{zeira2022paste,liu2023paste2,jones2023gpsa,clifton2023stalign},
building on Procrustes analysis~\citep{gower1975generalized,umeyama1991least}.
In each case the frame is fixed outside the dynamics.

\paragraph{Equivariance and quotient geometry.}
Equivariant generative models constrain how outputs transform when the
input frame
changes~\citep{kohler2020equivariant,satorras2021egnn,hoogeboom2022equivariant,klein2023equivariant}.
Quotient-space methods identify states related by a
symmetry, as in shape spaces, quotient diffusion, flow matching on
permutation orbits, and \rev{rigid-motion-invariant}
transport~\citep{kendall1984shape,xu2026quotient,wang2026ogpp,toukam2026sod}.
\citet{toukam2026sod} derive a Procrustes--Wasserstein distance from a
Benamou--Brenier action between probability measures \rev{that is invariant under rigid motions}.
Section~\ref{sec:theory} establishes the equivalence between the dynamic
and static formulations of unbalanced WFR transport with molecular state
(Theorem~\ref{thm:dynamic-static}), whose solution supplies the training targets of Section~\ref{sec:qfm}.
\rev{In their balanced setting the optimal translation matches the means,
and an orthogonal projection removes rigid velocity fields.  Under WFR the
optimal translation does not reduce to matching the means, and moving the
target far away has a finite cost, so an optimal pose needs a separate
existence argument (Appendix~\ref{app:equivalence}), here for a group that
moves positions but not molecular states.  We also give the projector in closed form for
finite measures with molecular state, show that optimally aligned paths are
horizontal (Lemma~\ref{lem:horizontal}) and that projected regression
recovers them (Theorem~\ref{thm:cfm}), and use one projector in training and
generation.  From WFR-FM we reuse the closed-form bridge between two matched
cells~\citep{liero2018oet,peng2026wfrfm}.}  Equivariance makes the
model consistent across frames, but an equivariant model can still predict a
shared rigid motion, which the projector removes.

\section{Tissue dynamics in a moving coordinate frame}
\label{sec:theory}

As summarized in Figure~\ref{fig:overview}, the method aligns consecutive
snapshots and matches their cells (panel b), follows closed-form WFR bridges
between matched cells to obtain targets (panel c), and regresses these
targets while ignoring shared rigid motion, which is also removed during
generation (panel d).  Below we derive
this projector from the WFR action and reduce the resulting dynamic
problem to endpoint alignment (see Table~\ref{tab:notation} in Appendix~\ref{app:notation} for a summary of notation).

\subsection{Evolving tissue measures}

A cell state $z=(x,q)$ pairs a position $x\in\R^{d_x}$ with a molecular
representation $q\in\R^{d_q}$.  A snapshot is the weighted point cloud
$\mu=\sum_i a_i\delta_{(x_i,q_i)}$, where $a_i>0$ is the abundance
represented by state $i$.  A tissue is thus a finite nonnegative measure on
$\cZ=\R^{d_x}\times\R^{d_q}$, and its total mass is its total abundance.  We
compare states through
$d_\Lambda^2((x,q),(x',q'))=\lambda_x\norm{x-x'}^2+\lambda_q\norm{q-q'}^2$,
where $\lambda_x,\lambda_q>0$ weight spatial and molecular displacement.

Let $\rho_t$ denote the tissue at time $t$, $u=(u_x,u_q)$ its velocity in
position and molecular state, and $g$ its relative growth rate.  The
continuity--reaction equation balances transport against changes in
abundance.  On an interval rescaled to $[0,1]$, the WFR distance selects the
fields joining $\rho_0=\mu_0$ to $\rho_1=\mu_1$ with least action:
\begin{align}
 \WFR_\delta^2(\mu_0,\mu_1)
 =\inf_{\rho,u,g}\frac12\int_0^1\!\int_{\cZ}
 &\left(\lambda_x\norm{u_x}^2+\lambda_q\norm{u_q}^2
 +\delta^2g^2\right)\dd\rho_t\dd t, \label{eq:wfr-action}\\[-2mm]
 &\partial_t\rho+\nabla_x\!\cdot(\rho u_x)
 +\nabla_q\!\cdot(\rho u_q)=g\rho , \nonumber
\end{align}
where $\delta>0$ controls the cost of mass change.  This action penalizes
every spatial displacement, including a translation or rotation of the whole
tissue, which is exactly what a change of coordinate frame between two
sections produces.

\subsection{Removing frame motion from the action}

The \rev{special Euclidean} group $G=\SE(d_x)$ of proper rotations and translations \rev{describes how a whole section can be turned and shifted.  It} acts on position
only, $h\cdot(x,q)=(Rx+b,q)$ for $h=(R,b)$.  The pushforward $h\push\mu$
moves every cell of $\mu$ by the same rigid motion.  We write
$[\mu]:=\{h_\#\mu:h\in G\}$ for the orbit of $\mu$, \rev{that is, the same tissue seen in every possible frame.  All its members have} the same relative spatial geometry, scale,
chirality, molecular states, and weights.
A moving rigid frame has velocity $a_t+\Omega_tx$ with
$\Omega_t^\top=-\Omega_t$\rev{, where $a_t$ is a common translation velocity and
$\Omega_t$ a rotation rate}.  We add this field to the spatial transport in
Equation~\eqref{eq:wfr-action} without charging for it:
\begin{align}
 \overline{\cA}_\delta(\mu_0,\mu_1)=
 \inf_{\rho,u,g,a,\Omega}\frac12\int_0^1\!\int_{\rev{\cZ}}
 &\left(\lambda_x\norm{u_x}^2+\lambda_q\norm{u_q}^2
 +\delta^2g^2\right)\dd\rho_t\dd t, \label{eq:qaction}\\[-1mm]
 \text{s.t.}\quad
 \partial_t\rho+\nabla_x\!\cdot\{\rho(u_x+a+\Omega x)\}
 &+\nabla_q\!\cdot(\rho u_q)=g\rho,
 \quad \rho_0=\mu_0,\ \rho_1=\mu_1 . \nonumber
\end{align}
Only motion relative to the frame now contributes to the spatial cost,
while molecular velocity and growth are penalized as before.
\rev{For example, if the second section is the first one turned by
$30^\circ$ about its centroid, Equation~\eqref{eq:wfr-action} pays for
moving every cell to its rotated position, most for the cells farthest from the centroid,
although nothing has changed in the tissue.  In
Equation~\eqref{eq:qaction} the frame term $a+\Omega x$ takes up this rotation
at no cost, so $u_x=0$ and the action vanishes.  If one region also
expands, only the expansion is charged.}

\subsection{Horizontal velocity and its projector}
\label{sec:projector}

Fix the total spatial velocity $v=u_x+a+\Omega x$ at one time.  The action
penalizes only $u_x=v-a-\Omega x$, so the optimal frame minimizes
$\int\norm{v-a-\Omega x}^2\dd\rho$ over translations $a$ and skew matrices
$\Omega$.  This is a weighted least-squares fit of a rigid field to $v$.  The
residual of the fit \rev{is the motion of cells relative to the tissue.  It} is orthogonal to every rigid field in $L^2(\rho)$, and we
call it the horizontal part of $v$.  The fit depends on the mass $M=\abs{\rho}$, the
centroid $\bar x=M^{-1}\!\int x\dd\rho$, the mean velocity
$\bar v=M^{-1}\!\int v\dd\rho$, the inertia
$S=\int(x-\bar x)(x-\bar x)^\top\dd\rho$, and the cross-moment
$C=\int(v-\bar v)(x-\bar x)^\top\dd\rho$.
\begin{proposition}[Closed-form horizontal projection]
\label{prop:projector}
The least-squares rigid component of $v$ is $a^\star=\bar v-\Omega^\star\bar
x$, where the minimum-norm skew solution satisfies
$\Omega^\star S+S\Omega^\star=C-C^\top$.  Consequently
\begin{equation}
 P_\rho^Hv(x,q)=v(x,q)-\bar v-\Omega^\star(x-\bar x)
 \label{eq:projector}
\end{equation}
is orthogonal in $L^2(\rho)$ to every rigid velocity field, and molecular
velocity and growth are unaffected.
\end{proposition}
In the planar case of our experiments, $\Omega^\star=\omega^\star J$, where
$J$ is the rotation by a right angle and
$\omega^\star=\int(x-\bar x)^\top J^\top(v-\bar v)\dd\rho\,\big/\!\int\norm{x-\bar x}^2\dd\rho$
is the angular momentum about the centroid divided by the moment of inertia
(Appendix~\ref{app:projection-details}).  \rev{Geometrically, the projector
subtracts from every cell the abundance-weighted mean velocity of the tissue and the rotation
about its centroid that best explains the velocities.}  Substituting the optimal frame into
Equation~\eqref{eq:qaction} leaves the ordinary continuity equation in $v$
and replaces the spatial kinetic term by $\lambda_x\norm{P^H_{\rho_t}v}^2$.  A
path whose spatial velocity is horizontal at every time is called a
\emph{horizontal lift} of the corresponding path of tissue shapes\rev{,
a path that changes the tissue without ever moving it as a whole}.

\subsection{An equivalent endpoint problem}
\label{sec:endpoint}

Solving Equation~\eqref{eq:qaction} directly requires optimizing over paths,
but the action also has a static form.  Define the distance between endpoint
orbits by
\begin{equation}
 \qWFR_\delta([\mu],[\nu])
 :=\inf_{h\in G}\WFR_\delta(\mu,h\push\nu).
 \label{eq:qwfr}
\end{equation}
A single pose acts on the whole target measure, so optimizing $h$ cannot
move cells individually.  Because $G$ acts by isometries \rev{(rigid motions preserve all distances)}, the value does not depend on which representatives are chosen.

\begin{theorem}[The equivalence between the dynamic and static formulations]
\leavevmode\phantomsection\label{thm:dynamic-static}
For nonzero finite atomic measures on $\cZ$ and $G=\SE(d_x)$, the infimum in
Equation~\eqref{eq:qwfr} is attained, $\qWFR_\delta$ is a metric on the orbit
space, and
$\overline{\cA}_\delta(\mu_0,\mu_1)
 =\min_{h\in G}\WFR_\delta^2(\mu_0,h\push\mu_1)
 =\qWFR_\delta^2([\mu_0],[\mu_1])$.
\end{theorem}

\rev{Ignoring frame motion therefore costs the same as ordinary WFR once the second section is placed in its best pose.}
The action identity follows from two bounds (Appendix~\ref{app:equivalence}).
\rev{Seen from its moving frame}, any admissible path becomes an ordinary WFR
path of the same cost ending at a \rev{rigidly moved copy of $\mu_1$}, which gives
the lower bound.  Conversely, a rigid frame at zero cost carries the WFR
geodesic to the optimally posed target $h^\star\push\mu_1$ back to $\mu_1$,
which gives the upper bound.

\section{Horizontal flow matching}
\label{sec:qfm}

Theorem~\ref{thm:dynamic-static} reduces the dynamic quotient problem to
endpoint alignment, from which we construct explicit bridge targets and a
projected regression loss.

\subsection{Joint alignment}

Alignment and correspondence depend on each other.  The relative pose
determines which cells are close, and the correspondence determines how the
sections should be aligned, so we estimate the two together.  For snapshots
$\mu=\sum_i a_i\delta_{z_i}$ and $\nu=\sum_j b_j\delta_{z'_j}$, let
$\gamma_{ij}\geq0$ be the mass transported from state $i$ to state $j$.
Quotient optimal entropy transport (qOET) solves
\begin{equation}
 (h^\star,\gamma^\star)\in\argmin_{h\in G,\,\gamma\geq0}
 2\delta^2\!\left[
 \ip{c_\delta(z_i,hz'_j)}{\gamma_{ij}}
 +\KL(\gamma\1\mid a)+\KL(\gamma^\top\1\mid b)
 \right],
 \label{eq:qoet}
\end{equation}
with $c_\delta(z,z')=-2\log\cos(d_\Lambda(z,z')/(2\delta))$ when
$d_\Lambda(z,z')<\pi\delta$ and $c_\delta=\infty$ otherwise\rev{, so states farther
apart than $\pi\delta$ are never matched}.  The first term
is the transport cost after alignment, and the two Kullback--Leibler terms
let the marginals of the coupling deviate from the observed abundances,
which accounts for the creation and destruction of
mass~\citep{liero2018oet,chizat2018scaling}.  At a fixed pose, the
unregularized minimum equals
$\WFR_\delta^2(\mu,h\push\nu)$~\citep{liero2018oet}.  Minimizing over the
pose as well gives $\qWFR_\delta^2$ and hence, by
Theorem~\ref{thm:dynamic-static}, the dynamic action.  A single pose is
shared by all pairs, whereas the coupling may split the mass of one cell
among several destinations.

\subsection{Bridges and training targets}
\label{sec:conditional-paths}

The coupling pairs the endpoints but does not specify the intermediate
states.  Along the coupling, the WFR geodesic between two measures splits
into independent paths of single weighted points, which we call bridges, so
it suffices to describe one pair.  Write $\gamma=\gamma^\star$.  A pair $(i,j)$ with
$\gamma_{ij}>0$ receives a source mass and a target mass,
\begin{equation}
 m_0=a_i\frac{\gamma_{ij}}{(\gamma\1)_i},\qquad
 m_1=b_j\frac{\gamma_{ij}}{(\gamma^\top\1)_j},
 \label{eq:semicoupling}
\end{equation}
which differ when abundance changes along the pair.  Set $z_0=z_i$ and
$\bar z_1=h^\star z'_j$.  The bridge moves one weighted point, called a
carrier, from $z_0$ to $\bar z_1$ while its mass changes from $m_0$ to $m_1$.
With $\vartheta=d_\Lambda(z_0,\bar z_1)/(2\delta)<\pi/2$ the half-angle
between the two endpoints, the WFR geodesic and its derivatives are
\begin{align}
 m_t&=(1-t)^2m_0+t^2m_1+2t(1-t)\sqrt{m_0m_1}\cos\vartheta,
 \qquad g_t^c=\dot m_t/m_t, \label{eq:mass-path}\\[-1mm]
 \alpha_t&=\vartheta^{-1}\atanTwo\!\left(t\sqrt{m_1}\sin\vartheta,\,
 (1-t)\sqrt{m_0}+t\sqrt{m_1}\cos\vartheta\right),\nonumber\\[-1mm]
 z_t&=(1-\alpha_t)z_0+\alpha_t\bar z_1,\qquad
 u_t^c=\frac{\sqrt{m_0m_1}\sin\vartheta}{\vartheta m_t}(\bar z_1-z_0). \label{eq:teacher}
\end{align}
The carrier moves along the segment from $z_0$ to $\bar z_1$ at a speed that
depends on its mass change, and $\alpha_t$ is the fraction of the segment
covered at time $t$.  The velocity $u_t^c=\dot z_t$ and the growth rate
$g_t^c$ are available in closed form, so no finite differences are needed
(Appendix~\ref{app:bridges} treats $\vartheta=0$).  A source state that is
not transported, $(\gamma\1)_i=0$, decays in place with mass $(1-t)^2a_i$,
and an untransported target state, $(\gamma^\top\1)_j=0$, appears at its
aligned position with mass $t^2b_j$.  Summing all components gives the
intermediate tissue at time $t$, which interpolates between $\mu$ and
$h^\star\push\nu$.

\subsection{Regressing velocity and growth}

The network predicts $(v_{\theta,x},v_{\theta,q},g_\phi)[\rho](z,t)$ from
the current tissue and time.  A training step draws one interval and one
local time $t$, then samples bridge states $z_i=(x_i,q_i)$ with masses
$m_i$, velocities $u_i=(u_i^x,u_i^q)$, and growth rates $g_i$ from the
intermediate tissue at that time.  Because all sampled states belong to the
same tissue, we fit one rigid field to all of their spatial errors
$e_i=v_{\theta,x}(z_i,t)-u_i^x$,
$(a_B,\Omega_B)=\argmin_{a,\Omega^\top=-\Omega}\sum_i
m_i\norm{e_i-a-\Omega x_i}^2$, and define the loss as
\begin{align}
 \cL_Q=\E\sum_i m_i\big[&\lambda_x
 \norm{e_i-a_B-\Omega_Bx_i}^2
 +\lambda_q\norm{v_{\theta,q}(z_i,t)-u^q_i}^2
 +\delta^2\abs{g_\phi(z_i,t)-g_i}^2\big].
 \label{eq:qfm-loss}
\end{align}
The spatial term applies the horizontal norm of Section~\ref{sec:projector}
to the regression residual, while molecular velocity and growth are fitted
directly.  The rigid field must be shared across cells, because a separate
rigid field per cell could absorb any residual and would remove the spatial
supervision entirely.  The next results explain why regressing
sampled bridges recovers the intended dynamics even when several bridges
pass through the same state.

\subsection{Population consistency}
\label{sec:guarantees}

Optimal bridges have a horizontal aggregate velocity (Lemma~\ref{lem:horizontal}),
and the projected regression loss recovers this aggregate
(Theorem~\ref{thm:cfm}).  Proofs are in Appendices~\ref{app:horizontal-proof}
and~\ref{app:cfm-proof}, respectively.

Several bridges may pass through one state at time $t$, and the aggregate
field averages their velocities weighted by current mass.  For an exact qOET
solution, this field contains no shared rigid motion.
\begin{lemma}[Optimally aligned geodesics are horizontal]
\label{lem:horizontal}
Let $h^\star$ attain Equation~\eqref{eq:qwfr} and let $\widetilde\rho_t$ be the
WFR geodesic from $\mu_0$ to $h^\star\push\mu_1$ obtained by marginalizing the
bridges of an exact qOET solution, with spatial aggregate velocity
$\widetilde u_x$.  Then $\widetilde u_x=P^H_{\widetilde\rho_t}\widetilde u_x$
in $L^2(\widetilde\rho_t)$ for almost every $t$, so $\widetilde\rho_t$
is a horizontal lift of a quotient geodesic.
\end{lemma}
\rev{The ideal training target thus never moves the tissue as a whole.}
Intuitively, removing a nonzero rigid component would lower the action
without changing the endpoint shapes, which would contradict optimality.  The
projector therefore leaves the ideal target unchanged.  A field learned from
finite data may still contain a rigid component, and since the loss in
Equation~\eqref{eq:qfm-loss} does not penalize it, we apply the projector
during generation to remove it.

In ordinary conditional flow matching, regression averages the conflicting
bridge velocities that pass through a
state~\citep{lipman2023flow,tong2023minibatch}.  The following identity
shows that this remains true when spatial errors are measured up to a rigid
field.  Here $Z_t$ is a bridge state sampled
according to mass, $\mathfrak{se}(d_x)$ denotes the rigid fields
$a+\Omega x$, and a tissue is regular when it has positive mass and no
nonzero rigid field vanishes on its support\rev{, which in the plane holds unless all
cells lie at one point}.
\begin{theorem}[Horizontal conditional flow matching identity]
\label{thm:cfm}
Fix $t$, let $U^c$ be the conditional bridge velocity with aggregate
$\bar U(z)=\E[U^c\mid Z_t=z]$, and assume $\rho_t$ is regular.  For every
candidate spatial field $v$,
\begin{equation}
 \inf_{\xi\in\mathfrak{se}(d_x)}
 \E\norm{v(Z_t)-U^c-\xi(Z_t)}^2
 =\inf_{\xi\in\mathfrak{se}(d_x)}
 \E\norm{v(Z_t)-\bar U(Z_t)-\xi(Z_t)}^2
 +C_t, \label{eq:cfm-identity}
\end{equation}
where $C_t$ does not depend on $v$ or $\xi$.  Hence every population
minimizer satisfies $P_{\rho_t}^Hv=P_{\rho_t}^H\bar U$.
\end{theorem}
\rev{Regressing each bridge velocity up to a shared rigid field has
the same minimizer as regressing their average, so bridges that cross do not
bias the learned field.}
Together, these results identify the qOET geodesic's horizontal and
molecular velocities and growth under exact population regression.
Integration recovers this path when the continuity equation is well posed
(Corollary~\ref{cor:recovery}), and adjacent optimal paths concatenate
(Proposition~\ref{prop:multitime}).  The guarantees require a global qOET
optimum, regular tissues, and exact regression over all fields.  The
implementation uses nonconvex pose search, sampled batches, and a fixed
number of carriers (Appendix~\ref{app:deferred}).

\section{The HORIZON model}
\label{sec:method}

\method{} solves qOET once per interval, trains a transformer on the
resulting bridges, and integrates the learned fields from the first
snapshot (Algorithm~\ref{alg:train}).

\paragraph{Computing supervision.}
We solve Equation~\eqref{eq:qoet} by alternating between the pose and the
coupling.  For a fixed pose, a column-generation solver works on a sparse set of
candidate pairs and adds improving pairs from the rest until the relative
primal--dual gap over the full support falls below $10^{-4}$, guaranteeing
optimality of the coupling within tolerance without storing the full cost matrix.  Because the
joint problem is nonconvex, we initialize from several starting poses and accept a
pose update only if it lowers the objective after re-solving the coupling.
This stopping criterion guarantees optimality of the coupling at the selected pose, while multi-start stability of the pose search itself is analyzed in
Appendix~\ref{app:qoet-init} (with solver details in Appendix~\ref{app:implementation}).

\paragraph{Predicting from the current tissue.}
The network is a transformer over the unordered set of cells in the current
tissue~\citep{vaswani2017attention,lee2019set}.  Each cell is embedded from
its molecular state, time, and position relative to the tissue moments, not
its own weight.  Each
pair of cells adds two invariant features to the attention logits, namely the dot
product of their centered positions and the signed area these positions
span.  Residual attention blocks update the embeddings, so every prediction
depends on the surrounding population.  Adding $\log m_j$ to the attention
logit of cell $j$ weights its contribution by the abundance it represents.
Splitting a weight among coincident copies therefore leaves the output
unchanged, so predictions depend only on the tissue measure. Since distances, dot products, and signed areas are unchanged by translation and
proper rotation, all hidden features are invariant.  The spatial head
outputs invariant coefficients that multiply a basis of directions attached
to the tissue, consisting of the centered position $x-\bar x$, its rotation $J(x-\bar x)$,
the covariance-transformed direction $\Sigma(x-\bar x)$, and pairwise
differences aggregated over the population.  Every basis vector rotates with
the tissue, so the predicted spatial velocity rotates as well.  Separate
heads output molecular velocity and growth, which are invariant
(Proposition~\ref{prop:architecture}).  The construction follows equivariant
point-cloud networks~\citep{satorras2021egnn}, and equivariance lets bridges
from different source frames train a single model.

\paragraph{Training.}
Each update passes a batch of weighted bridge states from one interval and
one time through the transformer and evaluates Equation~\eqref{eq:qfm-loss}.
Attention and the batch projector both operate on this sampled population.
On the real datasets we train for 9,000 Adam~\citep{kingma2015adam} updates
(Table~\ref{tab:hparams}).

\paragraph{Generating a trajectory.}
Generation starts from all cells of the initial snapshot.  At every RK4
stage, the network evaluates the current tissue, and the projector is
recomputed from its mass, centroid, and inertia.  Positions then follow
$P^H_\rho v_{\theta,x}$, molecular states follow $v_{\theta,q}$, and log
weights follow $g_\phi$, with a maximum step of 0.25 in normalized time.  The
number of carriers stays fixed, and generation uses neither endpoint
correspondences nor restarts from later observations.
\rev{Because unbalanced transport along coupled pairs often splits one source cell across multiple targets, a fixed carrier count can limit spatial coverage during rapid tissue expansion, an effect that can be mitigated at inference time by initializing multiple jittered carriers per cell without retraining (Appendix~\ref{app:support}).}

\section{Experiments}
\label{sec:experiments}

\paragraph{Data.}
We use axolotl brain regeneration, chicken heart development, \rev{serial
sections of one E14--16\,h Drosophila embryo indexed by depth, which plays the
role of time below,} and zebrafish embryogenesis~\citep{wei2022axolotl,mantri2021chicken,qiu2024holograms,liu2022zebrafish},
each with four or five snapshots and 3,550--46,209 cells in total.
Expression is represented by 32 principal components fitted on each series.
Coordinates keep their recorded orientation and origin, and mass is the cell
count relative to the initial snapshot.  All retained cells are used in qOET
and in evaluation, and the architecture and optimizer settings are the same
for all datasets (Table~\ref{tab:hparams}).

\paragraph{Baselines and protocols.}
We compare with WFR-FM, stVCR, STORIES, and NicheFlow on the same processed
inputs.  \method{} and the first three baselines generate the whole
trajectory continuously from the initial snapshot, whereas NicheFlow restarts each
interval from its observed source snapshot.  We therefore
also evaluate \method{} under this interval-restarted protocol, using the same trained model.
Because restarts provide observed intermediate states, Table~\ref{tab:real-data}
reports the two protocols separately (Appendix~\ref{app:baseline-details}).
STORIES predicts only molecular states, and NicheFlow does not predict mass
and is run on both raw and qOET-aligned coordinates.

\paragraph{Metrics.}
Sliced Wasserstein distance (SW)~\citep{bonneel2015sliced} and maximum mean
discrepancy (MMD)~\citep{gretton2012kernel} compare molecular and joint
spatial--molecular distributions, spatial SW and symmetric Chamfer distance
compare geometry, and relative error (RE) measures total abundance.  We use
two kinds of spatial scores to separate the effects of alignment and
projection.  \emph{Quotient} metrics first align each prediction to its
target by a proper $\SE(2)$ pose~\citep{umeyama1991least}, using the same
evaluator for every method, and thus measure error in tissue shape.
\emph{Direct} metrics skip this alignment and so also penalize rigid drift.
Because they require the intrinsic frame, they are reported only in
simulation\rev{, except in Table~\ref{tab:projector-real}, where real targets
are registered to the first snapshot by the composed qOET poses}.

\subsection{Robustness to changes of coordinate frame}
\label{sec:simulation}

\begin{table*}[t]
\centering\small
\caption{\textbf{Simulated tissue under hidden changes of coordinate frame.}  Mean $\pm$ standard error over completed fits ($n$ = completed/attempted, severe regime).  \emph{Direct} spatial SW is scored against intrinsic ground truth without alignment, measuring sensitivity to the frames used for training, whereas \emph{quotient} metrics are computed after $\SE(2)$ alignment.  The indented rows use the same network and training settings as \method{}.  \emph{Registered, no projector} aligns the snapshots in advance with the qOET poses and then trains and integrates without the projector, \emph{no quotient} does the same on the raw snapshots, and \emph{no growth} sets the reaction field to zero.}
\label{tab:simulation-ladder}
\setlength{\tabcolsep}{4pt}\renewcommand{\arraystretch}{0.92}
\resizebox{\textwidth}{!}{%
\begin{tabular}{lcrrrrrr}
\toprule
 & & \multicolumn{2}{c}{Direct spatial SW} & \multicolumn{4}{c}{Severe regime} \\
\cmidrule(lr){3-4}\cmidrule(lr){5-8}
Method & $n$ & Canonical & Severe & Quotient sp.\ SW & Quotient joint SW & Gene SW & Mass RE (\%) \\
\midrule
\method{} (ours) & 9/9 & $0.111\!\pm\!0.004$ & $0.115\!\pm\!0.003$ & $0.042\!\pm\!0.002$ & $0.037\!\pm\!0.001$ & $0.070\!\pm\!0.005$ & $0.02\!\pm\!0.01$ \\
\quad registered, no projector & 9/9 & -- & $0.127\!\pm\!0.004$ & $0.039\!\pm\!0.002$ & $0.034\!\pm\!0.001$ & $0.073\!\pm\!0.007$ & $0.02\!\pm\!0.01$ \\
\quad no quotient (raw, no projector) & 9/9 & -- & $1.246\!\pm\!0.093$ & $0.113\!\pm\!0.017$ & $0.121\!\pm\!0.039$ & $0.195\!\pm\!0.065$ & $0.23\!\pm\!0.04$ \\
\quad no growth (reaction set to zero) & 9/9 & -- & $0.117\!\pm\!0.003$ & $0.045\!\pm\!0.003$ & $0.039\!\pm\!0.002$ & $0.085\!\pm\!0.004$ & $9.32\!\pm\!0.00$ \\
\midrule
WFR-FM & 9/9 & $0.223\!\pm\!0.165$ & $1.213\!\pm\!0.027$ & $0.110\!\pm\!0.008$ & $0.142\!\pm\!0.016$ & $0.298\!\pm\!0.028$ & $0.14\!\pm\!0.03$ \\
stVCR & 9/9 & $0.115\!\pm\!0.015$ & $0.249\!\pm\!0.021$ & $0.119\!\pm\!0.006$ & $0.100\!\pm\!0.005$ & $0.160\!\pm\!0.012$ & $2.18\!\pm\!0.20$ \\
\bottomrule
\end{tabular}}
\end{table*}

\begin{table*}[t]
\centering\small
\caption{\textbf{Reconstruction of real spatial series.} We evaluate full-trajectory generation from the initial snapshot (\emph{first}) and interval-wise generation restarted from each observed source snapshot (\emph{restarted}). All scores average targets after the initial snapshot under proper $\SE(2)$ evaluation alignment, where lower values are better and boldface marks the best entry under each protocol. \method{} (first) and WFR-FM average three seeds, other rows use one seed, and \method{} (restarted) evaluates one trained model (seed 17). Dashes mark outputs not predicted or scored (Appendix~\ref{app:baseline-details}).}
\label{tab:real-data}
\setlength{\tabcolsep}{4pt}\renewcommand{\arraystretch}{0.9}
\resizebox{\textwidth}{!}{%
\begin{tabular}{lllrrrrrrr}
\toprule
Dataset & Method & Protocol & Joint SW & Joint MMD & Spatial SW & Chamfer & Gene SW & Gene MMD & Mass RE (\%) \\
\midrule
 \multirow{7}{*}{Axolotl} & \method{} & first & \best{0.063} & \best{0.003} & \best{0.240} & \best{0.367} & \best{0.122} & \best{0.018} & \best{0.09} \\
  & WFR-FM & first & 1.441 & 0.167 & 5.907 & 2.860 & 1.171 & 0.214 & 146.3 \\
  & stVCR & first & 1.040 & 0.218 & 4.294 & 1.008 & 0.248 & 0.044 & 1.30 \\
  & STORIES & first & -- & -- & -- & -- & 1.087 & 0.207 & -- \\
\cmidrule(l){2-10}
  & \method{} & restarted & \best{0.054} & \best{0.001} & \best{0.224} & \best{0.323} & \best{0.046} & \best{0.006} & -- \\
  & NicheFlow raw & restarted & 0.201 & 0.025 & 0.843 & 0.360 & 0.279 & 0.060 & -- \\
  & NicheFlow aligned & restarted & 0.270 & 0.038 & 1.055 & 0.383 & 0.345 & 0.076 & -- \\
\midrule
 \multirow{7}{*}{Chicken} & \method{} & first & \best{0.134} & \best{0.037} & \best{0.199} & \best{0.392} & \best{0.597} & \best{0.098} & \best{0.45} \\
  & WFR-FM & first & 0.187 & 0.064 & 0.540 & 0.576 & 0.751 & 0.135 & 1.56 \\
  & stVCR & first & 0.576 & 0.205 & 1.806 & 2.079 & 1.518 & 0.262 & 99.2 \\
  & STORIES & first & -- & -- & -- & -- & 1.476 & 0.271 & -- \\
\cmidrule(l){2-10}
  & \method{} & restarted & \best{0.061} & \best{0.014} & \best{0.081} & 0.224 & 0.252 & \best{0.034} & -- \\
  & NicheFlow raw & restarted & 0.064 & 0.025 & 0.198 & 0.150 & \best{0.217} & 0.042 & -- \\
  & NicheFlow aligned & restarted & 0.079 & 0.035 & 0.268 & \best{0.149} & 0.231 & 0.047 & -- \\
\midrule
 \multirow{7}{*}{Drosophila} & \method{} & first & \best{0.035} & \best{0.013} & \best{0.043} & \best{0.125} & \best{0.107} & \best{0.023} & \best{0.10} \\
  & WFR-FM & first & 0.074 & 0.031 & 0.064 & 0.152 & 0.157 & 0.032 & 0.23 \\
  & stVCR & first & 0.228 & 0.112 & 0.836 & 0.669 & 0.427 & 0.110 & 11.5 \\
  & STORIES & first & -- & -- & -- & -- & 0.797 & 0.241 & -- \\
\cmidrule(l){2-10}
  & \method{} & restarted & \best{0.018} & \best{0.005} & \best{0.029} & \best{0.103} & \best{0.081} & \best{0.017} & -- \\
  & NicheFlow raw & restarted & 0.059 & 0.030 & 0.158 & 0.140 & 0.160 & 0.045 & -- \\
  & NicheFlow aligned & restarted & 0.088 & 0.046 & 0.185 & 0.142 & 0.175 & 0.051 & -- \\
\midrule
 \multirow{7}{*}{Zebrafish} & \method{} & first & \best{0.071} & \best{0.009} & \best{0.206} & \best{0.358} & \best{0.289} & \best{0.058} & \best{0.51} \\
  & WFR-FM & first & 1.035 & 0.160 & 4.049 & 3.142 & 1.441 & 0.374 & 45.0 \\
  & stVCR & first & 1.862 & 0.331 & 7.394 & 7.855 & 1.862 & 0.377 & 99.5 \\
  & STORIES & first & -- & -- & -- & -- & 0.831 & 0.215 & -- \\
\cmidrule(l){2-10}
  & \method{} & restarted & \best{0.028} & \best{0.003} & \best{0.060} & \best{0.184} & \best{0.108} & \best{0.023} & -- \\
  & NicheFlow raw & restarted & 0.135 & 0.029 & 0.537 & 0.264 & 0.243 & 0.058 & -- \\
  & NicheFlow aligned & restarted & 0.141 & 0.029 & 0.561 & 0.222 & 0.251 & 0.052 & -- \\
\bottomrule
\end{tabular}}
\end{table*}

A simulated tissue bends, contracts, and changes mass and molecular state.
Each of three intrinsic trajectories has six snapshots and is observed in one
canonical, three mildly perturbed, and three severely perturbed frames.  The
applied poses are hidden from training, and Table~\ref{tab:simulation-ladder}
scores the five snapshots after the first.

\paragraph{Sensitivity to the frame.}
As the perturbation grows from canonical to severe, the direct spatial SW of
\method{} rises only from 0.111 to 0.115, while that of WFR-FM grows more
than fivefold and that of stVCR more than doubles.  \method{} also has lower
quotient spatial and joint errors than both baselines, so its advantage
extends to tissue shape.

\paragraph{Contributions of alignment and projection.}
Aligning the snapshots in advance with the qOET poses but omitting the projector
gives a quotient spatial SW of 0.039.  This is close to the 0.042 of the full
model and far below the 0.113 obtained without either ingredient, so
alignment accounts for the gain in reconstructed shape.  The projector
targets rigid drift, which evaluation alignment removes, so its effect
appears in the direct metrics.  In a matched ablation retrained under identical seeds on the nine severe variants (Appendix~\ref{app:formal-simulation}), adding the projector lowers direct
spatial SW from 0.127 to 0.116 (paired reduction $0.011\pm0.003$, consistent with the $0.115$ full-model mean in Table~\ref{tab:simulation-ladder}) and reduces direct Chamfer across every variant by $0.013\pm0.003$, while the difference in
quotient spatial SW stays within one standard error of zero.  Setting growth
to zero raises mass RE from $0.02\%$ to $9.3\%$ with little change in spatial
error (Appendices~\ref{app:formal-simulation} and~\ref{app:sensitivity} give
the full benchmark and sensitivity).

\subsection{Reconstruction and interpolation of real tissues}
\label{sec:full-results}

\begin{figure*}[t]
\centering
\includegraphics[width=.64\textwidth]{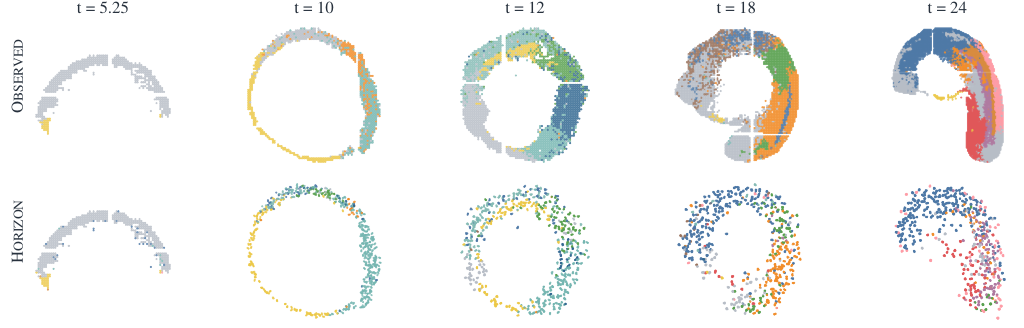}
\caption{\textbf{Reconstructing zebrafish development from the initial snapshot.}
Observations are shown above and a representative \method{} trajectory below, with time in hours
post fertilization.  Predictions use no restarts and are aligned for
evaluation.  Points carry equal mass so point count reflects abundance, while colors
show observed annotations above and annotations decoded from predicted
expression below (Appendix~\ref{app:visuals}).}
\label{fig:zebrafish-rollout}
\end{figure*}

\paragraph{Reconstruction from the initial snapshot.}
In Table~\ref{tab:real-data}, \method{} has lower mean error than WFR-FM and
stVCR on all seven metrics and all four datasets, and lower molecular error
than STORIES.  When restarted at each observed snapshot, it improves over both
NicheFlow variants across all metrics on axolotl, Drosophila, and zebrafish,
trailing only on chicken's Chamfer and gene SW.  Figure~\ref{fig:zebrafish-rollout} shows the
generated changes in organization and abundance, and mean mass RE stays
below $0.51\%$ in every series.  Spatial gains are largest on axolotl and
zebrafish and smallest on Drosophila, whose frames are nearly aligned.
An ablation without alignment and projection, which converges on two series,
raises spatial SW from 0.199 to 0.448 on chicken but only from 0.043 to 0.049
on Drosophila (Table~\ref{tab:quotient-off}).  \rev{Across all 15 intervals, our eight-start pose search reaches the same orientation as a 120-start search, with orientations over $20^\circ$ away worsening the objective by at least 1.6\% on axolotl, Drosophila, and zebrafish and at most 0.5\% on the nearly round early chicken section (Appendix~\ref{app:qoet-init}).  Moreover, because pure creation and decay carry at most 0.021\% of target mass whereas source splitting is widespread (Table~\ref{tab:teacher-stats}), initializing 16 jittered carriers per initial cell without retraining lowers chicken Chamfer distance from 0.389 to 0.210 (Appendix~\ref{app:support}).}

\begin{table}[t]
\centering\small
\caption{\rev{\textbf{Registration and projection on real data.} All rows use the same qOET bridges (written in the frame of the first snapshot for the ambient loss) and the same network and optimizer, averaged over seeds 17, 29, and 43.  Gene SW is frame-free, while direct spatial SW scores the generated trajectory against targets registered to the first frame by the composed qOET poses and thus also counts rigid drift.  Appendix~\ref{app:projector-real} gives all metrics and standard deviations.}}
\label{tab:projector-real}
\setlength{\tabcolsep}{3.2pt}
\resizebox{\columnwidth}{!}{%
\begin{tabular}{lccrrrrrrrr}
\toprule
 & \multicolumn{2}{c}{Projection} & \multicolumn{2}{c}{Axolotl} & \multicolumn{2}{c}{Chicken} & \multicolumn{2}{c}{Drosophila} & \multicolumn{2}{c}{Zebrafish} \\
\cmidrule(lr){2-3}\cmidrule(lr){4-5}\cmidrule(lr){6-7}\cmidrule(lr){8-9}\cmidrule(lr){10-11}
Variant & loss & gen. & Gene SW & Direct SW & Gene SW & Direct SW & Gene SW & Direct SW & Gene SW & Direct SW \\
\midrule
\method{} & \cmark & \cmark & \best{0.123} & 0.239 & \best{0.581} & 0.215 & \best{0.103} & \best{0.041} & \best{0.283} & \best{0.192} \\
registered, no projector &  &  & 0.126 & \best{0.228} & 0.641 & 0.260 & 0.107 & 0.043 & 0.289 & 0.208 \\
projection in loss only & \cmark &  & \best{0.123} & 0.549 & \best{0.581} & 0.243 & \best{0.103} & 0.281 & \best{0.283} & 0.741 \\
projection in generation only &  & \cmark & 0.126 & 0.231 & 0.641 & \best{0.213} & 0.107 & \best{0.041} & 0.289 & 0.201 \\
\bottomrule
\end{tabular}}
\end{table}

\rev{\paragraph{Registration and projection on real data.}
Table~\ref{tab:projector-real} separates the two on all four series.
Training on the same qOET bridges with the ambient loss and integrating
without the projector gives tissue shapes close to those of \method{} after
evaluation alignment, as Lemma~\ref{lem:horizontal} predicts, since a shared rigid
motion carries at most 0.42\% of the bridges' spatial kinetic energy on
every interval, against 77--98\% on chicken when the pose is fixed to the
identity (Table~\ref{tab:teacher-stats}).  At the same integrator, the
horizontal loss still lowers the mean gene SW on all four series, by 2--9\%,
and \method{} has lower direct spatial SW than this registered model on three
of the four.  The horizontal loss leaves the rigid part of the network output free (2--37\% of its spatial
kinetic energy), so the projector is needed in generation as well.  Without it, the generated trajectory rotates by up to $16.5^\circ$ and direct spatial SW grows up to
6.8-fold (Appendix~\ref{app:projector-real}).}

\paragraph{Interpolation at unobserved times.}
We hold out each interior snapshot before coupling and training and score
the omitted target.
\rev{All statistics pooled across sections, including the PCA and the spatial
scale, are refit without it.}
\method{} completes all 11 folds \rev{with three seeds, outperforming stVCR across nearly every fold and metric and achieving lower spatial SW, Chamfer, and held-out mass error than WFR-FM on the vast majority of folds, with the largest margins on axolotl and zebrafish where sections differ most in frame (Appendix~\ref{app:holdout}).}  A \method{} fit, including qOET, takes 12--20
minutes on one GPU, 2.6--11.4 times faster than stVCR and 20--30 times
faster than NicheFlow (Appendix~\ref{app:baseline-details}).  \rev{The joint pose and coupling search takes 6--15 minutes of this, and the projector adds little (Appendix~\ref{app:runtime-breakdown}).}

\section{Discussion and conclusion}
\label{sec:discussion}

Horizontal flow matching learns tissue dynamics consistently across coordinate
frames by combining joint alignment with horizontal projection. Alignment
provides frame-aware training paths, while the projector removes rigid motion
from both regression and generation. \rev{On real data, alignment accounts for
most of the gain in tissue shape, the horizontal loss further lowers molecular
error, and the shared projector keeps generation free of rigid drift.} The resulting fields jointly describe
tissue organization, molecular state, and abundance, and provide readouts of
biological programs and coupling affinities
(Appendix~\ref{app:biology}).

\textbf{Limitations.}
The quotient formulation removes any collective rigid motion that is
indistinguishable from a change of coordinate frame, so genuinely rigid
tissue motion cannot be recovered from these observations alone. In addition,
the current model is fitted separately to each time course, leaving
generalization across specimens to future work. \rev{Generation uses a fixed
set of carriers, so mass splitting appears only as the spread of carriers.  Splitting carriers during integration would remove the need to smooth the
initial measure.  When a section is small and nearly round, as chicken D4,
several poses are almost equally good, and the pose is then weakly
identified by the snapshots themselves.}
Finally, scalability remains
a practical limitation, since both coupling construction and attention have
quadratic complexity in the number of cells.
\label{page:concend}

\FloatBarrier
\label{page:endofbody}
\section*{AI Use Statement}

The authors defined the research objective and core idea.  Part of the
algorithmic design was assisted by our agentic system
\mbox{CellCompass}~\citep{zhang2026cellcompass}, and generative AI tools were
used to assist with the mathematical derivations, implementation refinement,
numerical experiments, figure preparation, and manuscript editing.  We reviewed all AI-assisted work.  The authors checked the
derivations and proofs step by step, verified AI-assisted code against
independent numerical checks, checked every figure against the underlying
results, and reviewed all edited text for correctness.  The authors
interpreted the results, verified the final claims, and take responsibility
for the final content of this work, including text, claims, and artifacts
produced with the aid of generative AI.

\section*{Reproducibility Statement}

Proofs of all theoretical results and their standing assumptions are provided
in Appendix~\ref{app:proofs}.
Appendix~\ref{app:implementation} connects these population guarantees to the
finite implementation and specifies the qOET solver, the network, training
and generation (Algorithm~\ref{alg:train}), preprocessing, and the
evaluation metrics, and Table~\ref{tab:hparams} lists the settings for each
dataset.  Appendix~\ref{app:formal-simulation} describes the
frame-perturbation benchmark, including the seeds that generate every view,
and Appendix~\ref{app:baseline-details} gives the baseline configurations,
evaluation protocols, and runtimes.  Appendices~\ref{app:sensitivity}
and~\ref{app:holdout} report the sensitivity and timepoint holdout studies.

\bibliography{references}
\bibliographystyle{iclr2027_conference}

\appendix
\raggedbottom
\clearpage

\numberwithin{theorem}{section}
\renewcommand{\theHtheorem}{\thesection.\arabic{theorem}}
\section{Proofs and supplementary results}
\label{app:proofs}

We use the notation of Sections~\ref{sec:theory}--\ref{sec:qfm}.
Continuity equations are understood in the weak sense, and admissible frame
coefficients are integrable in time.  Whenever a horizontal projection is
used, the measure has positive mass and finite spatial second moment, and
the projected field belongs to $L^2(\rho)$.  These conditions make the
weighted moments and the rigid-frame flow well defined.  The distinction
between the population results and their numerical implementation is given
in Appendix~\ref{app:deferred}.

\subsection{Proof of Proposition~\ref{prop:projector}}
\label{app:projection-details}

\textit{Let $\rho$ have mass $M>0$, and let $\bar x,\bar v,S,C$ be the
weighted moments defined in Section~\ref{sec:projector}.  The least-squares
rigid component of $v$ is $a^\star+\Omega^\star x$, where}
\[
 a^\star=\bar v-\Omega^\star\bar x,\qquad
 \Omega^\star S+S\Omega^\star=C-C^\top.
\]
\textit{The minimum-norm skew solution defines
$P_\rho^Hv=v-\bar v-\Omega^\star(x-\bar x)$, the orthogonal projection
onto the complement of the rigid fields in $L^2(\rho)$.}
\begin{proof}
Consider the convex quadratic objective
\[
 F(a,\Omega)=\int\norm{v-a-\Omega x}^2\dd\rho,
 \qquad \Omega^\top=-\Omega.
\]
For any translation perturbation $b$, stationarity gives
\[
 0=\left.\frac{\dd}{\dd\epsilon}F(a+\epsilon b,\Omega)
       \right|_{\epsilon=0}
 =-2b^\top\int(v-a-\Omega x)\dd\rho.
\]
Hence $a=\bar v-\Omega\bar x$.  Writing $y=x-\bar x$ and
$w=v-\bar v$ reduces the objective to
$F(\Omega)=\int\norm{w-\Omega y}^2\dd\rho$.
For any skew matrix $K$,
\begin{align*}
 0&=-\frac12\left.\frac{\dd}{\dd\epsilon}F(\Omega+\epsilon K)
       \right|_{\epsilon=0}\\
  &=\int(w-\Omega y)^\top Ky\dd\rho
   =\operatorname{tr}\big((C^\top+S\Omega)K\big).
\end{align*}
Orthogonality to all skew matrices means that $C^\top+S\Omega$ is
symmetric.  Using $S^\top=S$ and $\Omega^\top=-\Omega$ gives
\[
 C^\top+S\Omega=C-\Omega S,
 \qquad \Omega S+S\Omega=C-C^\top.
\]
These normal equations are sufficient because $F$ is convex.  Their
minimum-norm solution is given by the pseudoinverse of
$\mathcal L_S:\Omega\mapsto\Omega S+S\Omega$ on
$\mathfrak{so}(d_x)$.  If $S$ is positive definite, this operator is
invertible, since in an eigenbasis of $S$ its action on the $(i,j)$ entry is
multiplication by $s_i+s_j>0$.  In the singular case, any two minimizing
rigid fields agree $\rho$-almost everywhere, so the projected field remains
unique.

For the residual $r=v-a^\star-\Omega^\star x$, the two stationarity
conditions imply
\[
 \int r\dd\rho=0,\qquad
 \int r^\top Kx\dd\rho=0\quad(K^\top=-K).
\]
It follows that $\langle r,b+Kx\rangle_{L^2(\rho)}=0$ for every rigid
field, which proves the claimed orthogonal projection.

In two dimensions, write $\Omega^\star=\omega^\star J$ with $J$ the
quarter turn.  Differentiating $F(\omega J)$ gives
\[
 \omega^\star
 =\frac{\int(Jy)^\top w\dd\rho}{\int\norm{Jy}^2\dd\rho}
 =\frac{\int(x-\bar x)^\top J^\top(v-\bar v)\dd\rho}
        {\int\norm{x-\bar x}^2\dd\rho},
\]
when the denominator is positive, and the minimum-norm choice is zero when it
vanishes.  A measure is regular when no nonzero rigid field vanishes
$\rho$-almost everywhere.  In two dimensions, this is equivalent to
positive mass and a positive denominator in the preceding expression.
\end{proof}

\subsection{Proof of Theorem~\ref{thm:dynamic-static}}
\label{app:equivalence}

\textit{For nonzero finite atomic measures on $\cZ$ and $G=\SE(d_x)$,
the infimum in Equation~\eqref{eq:qwfr} is attained,
$\qWFR_\delta$ is a metric on the orbit space, and}
\[
 \overline{\cA}_\delta(\mu_0,\mu_1)
 =\min_{h\in G}\WFR_\delta^2(\mu_0,h\push\mu_1)
 =\qWFR_\delta^2([\mu_0],[\mu_1]).
\]
\begin{proof}
Write $d=\WFR_\delta$.  Since $G$ acts only on the spatial coordinates
and preserves $d_\Lambda$, it acts by isometries of $d$.

\emph{Existence of an optimal pose.}
Let $h_n=(R_n,b_n)$ be a minimizing sequence.  The rotations lie in the
compact group $SO(d_x)$.  If the translations are bounded, a subsequence
converges to $h^\star$, and continuity of WFR on finite atomic measures gives
\[
 d^2(\mu,h^\star\push\nu)=\inf_{h\in G}d^2(\mu,h\push\nu).
\]
If the translations are unbounded, pass to a subsequence with
$\norm{b_n}\to\infty$.  Both supports are finite, so all cross-distances
eventually exceed the transport cut locus $\pi\delta$.  The optimal
coupling then vanishes and
\[
 d^2(\mu,h_n\push\nu)=2\delta^2(|\mu|+|\nu|).
\]
If the infimum is smaller than this value, such a subsequence cannot be
minimizing.  If the infimum equals this value, any fixed translation that
separates the supports beyond the cut locus attains it.  Thus an optimizer
exists in both cases.

\emph{Metricity.}
For representatives changed by $g_0,g_1\in G$, isometry gives
\[
 d(g_0\push\mu,h\push g_1\push\nu)
 =d(\mu,(g_0^{-1}hg_1)\push\nu).
\]
The map $h\mapsto g_0^{-1}hg_1$ is a bijection of $G$, so the infimum
depends only on the orbits.  Nonnegativity follows from that of $d$, and
symmetry follows by replacing $h$ with $h^{-1}$.  For $\epsilon$-optimal
poses $h_1,h_2$ for $(\mu,\nu)$ and $(\nu,\eta)$,
\begin{align*}
 \qWFR_\delta([\mu],[\eta])
 &\le d(\mu,(h_1h_2)\push\eta)\\
 &\le d(\mu,h_1\push\nu)
      +d(h_1\push\nu,(h_1h_2)\push\eta)\\
 &=d(\mu,h_1\push\nu)+d(\nu,h_2\push\eta)\\
 &\le\qWFR_\delta([\mu],[\nu])
      +\qWFR_\delta([\nu],[\eta])+2\epsilon.
\end{align*}
Letting $\epsilon\downarrow0$ proves the triangle inequality.  Finally,
if the quotient distance is zero, attainment and metricity of $d$ imply
$\mu=h^\star\push\nu$, hence $[\mu]=[\nu]$.

\emph{Lower bound for the dynamic action.}
Take an admissible tuple $(\rho,u,g,a,\Omega)$ in
Equation~\eqref{eq:qaction}.  Let $G_t(x)=R_tx+b_t$ solve
\[
 \partial_tG_t(x)=a_t+\Omega_tG_t(x),\qquad G_0=I,
\]
and put $\widetilde\rho_t=(G_t^{-1})\push\rho_t$.
For $f\in C_c^1(\cZ)$, define $\psi_t(x,q)=f(G_t^{-1}x,q)$.
The chain rule gives
\[
 \partial_t\psi_t+\nabla_x\psi_t\cdot(a_t+\Omega_tx)=0.
\]
Testing the weak continuity equation against $\psi_t$ therefore yields
\begin{align*}
 \frac{\dd}{\dd t}\int f\dd\widetilde\rho_t
 &=\int\big[\partial_t\psi_t
   +\nabla_x\psi_t\cdot(u_x+a_t+\Omega_tx)
   +\nabla_q\psi_t\cdot u_q+g\psi_t\big]\dd\rho_t\\
 &=\int\big[\nabla_x f\cdot\widetilde u_x
   +\nabla_q f\cdot\widetilde u_q+\widetilde g f\big]
   \dd\widetilde\rho_t,
\end{align*}
where
\[
 \widetilde u_x(x,q,t)=R_t^\top u_x(G_tx,q,t),\quad
 \widetilde u_q(x,q,t)=u_q(G_tx,q,t),\quad
 \widetilde g(x,q,t)=g(G_tx,q,t).
\]
Thus the pulled-back fields satisfy the ordinary continuity--reaction
equation.  Since $R_t$ is orthogonal,
\begin{align*}
 &\int\big(\lambda_x\norm{\widetilde u_x}^2
       +\lambda_q\norm{\widetilde u_q}^2+\delta^2\widetilde g^2\big)
       \dd\widetilde\rho_t\\
 &\hspace{12mm}=\int\big(\lambda_x\norm{u_x}^2
       +\lambda_q\norm{u_q}^2+\delta^2g^2\big)\dd\rho_t.
\end{align*}
The pulled-back path has endpoints $\mu_0$ and
$(G_1^{-1})\push\mu_1$.  Its action is consequently at least
$d^2(\mu_0,(G_1^{-1})\push\mu_1)$, and taking the infimum over admissible
tuples gives
\[
 \overline{\cA}_\delta(\mu_0,\mu_1)
 \ge\min_{h\in G}d^2(\mu_0,h\push\mu_1).
\]

\emph{Upper bound.}
Fix $h\in G$ and an ambient WFR geodesic $\widetilde\rho_t$ joining
$\mu_0$ to $h\push\mu_1$.  Since $\SE(d_x)$ is connected, choose a
smooth rigid path $G_t$ from $I$ to $h^{-1}$ and set
$\rho_t=G_t\push\widetilde\rho_t$.  Define the relative fields by
\[
 u_x(G_tx,q,t)=R_t\widetilde u_x(x,q,t),\quad
 u_q(G_tx,q,t)=\widetilde u_q(x,q,t),\quad
 g(G_tx,q,t)=\widetilde g(x,q,t),
\]
and place $\dot G_t\circ G_t^{-1}=a_t+\Omega_tx$ in the frame term.
The preceding change of variables, read in reverse, proves feasibility.
Moreover, $\rho_0=\mu_0$, $\rho_1=\mu_1$, and the action equals
$d^2(\mu_0,h\push\mu_1)$.  Minimizing over $h$ gives the reverse inequality.
\end{proof}

\subsection{Frame covariance}
\label{app:frame-covariance}

\begin{proposition}[Covariance under changes of representatives]
\label{prop:frame-covariance}
For $g_0,g_1\in G$, changing the endpoint representatives to
$g_0\push\mu_0,g_1\push\mu_1$ sends every optimal relative pose $h$
to $g_0hg_1^{-1}$ and preserves its OET coupling.  In particular,
\begin{equation}
 \operatorname*{Argmin}_{h}\WFR_\delta
 (g_{0\#}\mu_0,h_{\#}g_{1\#}\mu_1)
 =g_0\operatorname*{Argmin}_{h}\WFR_\delta(\mu_0,h_{\#}\mu_1)g_1^{-1}.
 \label{eq:argmin-covariance}
\end{equation}
For a spatial field, with $(g_*v)(gx,q)=Rv(x,q)$ when $g=(R,b)$,
\[
 P^H_{g\push\rho}(g_*v)=g_*(P^H_\rho v).
\]
\end{proposition}
\begin{proof}
Ground-metric invariance gives, for every pair of endpoint states,
\[
 d_\Lambda(g_0z_i,(g_0hg_1^{-1})g_1z'_j)
 =d_\Lambda(z_i,hz'_j).
\]
All transport costs are unchanged, as are the weights and KL penalties.
The map $h\mapsto g_0hg_1^{-1}$ is bijective, which proves the assertion
about poses and couplings in both directions.

For the projector, the transformed moments are
\[
 \bar x'=R\bar x+b,\qquad \bar v'=R\bar v,\qquad
 S'=RSR^\top,\qquad C'=RCR^\top.
\]
The minimum-norm skew solution consequently transforms as
$\Omega'=R\Omega R^\top$, because it solves the transformed Sylvester equation
and conjugation preserves the Frobenius norm.  Substitution gives
\begin{align*}
 (P^H_{g\push\rho}g_*v)(gx,q)
 &=Rv(x,q)-R\bar v
        -(R\Omega R^\top)\big(Rx+b-(R\bar x+b)\big)\\
 &=R\big(v(x,q)-\bar v-\Omega(x-\bar x)\big).
\end{align*}
This is the claimed covariance.  The pose identity is set-valued and does
not require a unique optimizer.  For symmetric tissues, different
minimizers may give different representative-level paths, each optimal in
the quotient.  Selecting a unique equivariant lift would additionally
require an equivariant choice among those minimizers.
\end{proof}

\subsection{Analytic WFR bridges}
\label{app:bridges}

\begin{proposition}[Bridge construction and marginal dynamics]
\label{prop:bridge-construction}
Fix a pose $h$ and an optimal unregularized OET coupling $\gamma$ between
$\mu=\sum_i a_i\delta_{z_i}$ and $h\push\nu$.
For every positive edge, assign the two masses in
Equation~\eqref{eq:semicoupling} and use the bridge in
Equations~\eqref{eq:mass-path}--\eqref{eq:teacher}.  Add the pure decay and
creation components for zero rows and columns.  Their sum has endpoints
$\mu,h\push\nu$, and the mass-weighted aggregate velocity and growth
satisfy the continuity--reaction equation.  The resulting path is an
ambient WFR geodesic.
\end{proposition}
\begin{proof}
\emph{Endpoint masses.}
Write $r_i=\sum_j\gamma_{ij}$ and $s_j=\sum_i\gamma_{ij}$.  The OET
coupling has soft marginal constraints, so $r$ and $s$ need not equal $a$
and $b$.  Instead, the two masses assigned to each positive edge satisfy
\[
 \sum_j m_{0,ij}
 =\sum_j a_i\frac{\gamma_{ij}}{r_i}=a_i\quad(r_i>0),\qquad
 \sum_i m_{1,ij}
 =\sum_i b_j\frac{\gamma_{ij}}{s_j}=b_j\quad(s_j>0).
\]
A zero row contributes a decay component of initial mass $a_i$, and a zero
column contributes a creation component of terminal mass $b_j$.
Consequently, the full collection of components has the prescribed
endpoint measures.  It is this pair of mass allocations, or semi-coupling,
that restores the endpoints~\citep{liero2018oet,peng2026wfrfm}.

\emph{A single transported component.}
Let $z_0,z_1$ be its endpoints and let
$\vartheta=d_\Lambda(z_0,z_1)/(2\delta)<\pi/2$.
In the WFR cone, a weighted state has radial coordinate $\sqrt m$, and
the squared cone distance between these two states is
\[
 d_{\mathrm{cone}}^2=m_0+m_1-2\sqrt{m_0m_1}\cos\vartheta.
\]
The WFR action of the corresponding single-component geodesic is
$2\delta^2d_{\mathrm{cone}}^2$ under the convention of
Equation~\eqref{eq:wfr-action}.  The straight segment in the cone plane is
represented by
\[
 A_t=(1-t)\sqrt{m_0}+t\sqrt{m_1}e^{i\vartheta},\qquad
 m_t=|A_t|^2,\qquad \phi_t=\arg A_t.
\]
Expanding $|A_t|^2$ gives Equation~\eqref{eq:mass-path}.  Moreover,
\begin{align*}
 \dot\phi_t
 &=\frac{\operatorname{Im}(\overline{A_t}\dot A_t)}{|A_t|^2}
   =\frac{\sqrt{m_0m_1}\sin\vartheta}{m_t},\\
 \dot z_t
 &=\frac{\dot\phi_t}{\vartheta}(z_1-z_0)
   =\frac{\sqrt{m_0m_1}\sin\vartheta}
          {\vartheta m_t}(z_1-z_0).
\end{align*}
Here $z_t=(1-\phi_t/\vartheta)z_0+(\phi_t/\vartheta)z_1$.
Differentiating the mass gives
\begin{equation}
 \dot m_t=-2(1-t)m_0+2tm_1
          +2(1-2t)\sqrt{m_0m_1}\cos\vartheta,
 \qquad g_t^c=\dot m_t/m_t.
 \label{eq:bridge-mass-derivative}
\end{equation}
At $\vartheta=0$, the continuous limit is
\[
 \alpha_t=\frac{t\sqrt{m_1}}{(1-t)\sqrt{m_0}+t\sqrt{m_1}},
 \qquad \frac{\sin\vartheta}{\vartheta}=1.
\]
Pure decay and creation have masses $(1-t)^2m_0$ and $t^2m_1$ and
growth rates $-2/(1-t)$ and $2/t$, respectively.  The rates are used only
at positive-mass times.  Their weighted squared growth is $4m_0$ and
$4m_1$, so the singular zero-mass endpoint contributes no divergence to
the action.

\emph{The marginal continuity equation.}
Index all components by $\ell$ and put
$\rho_t=\sum_\ell m_\ell(t)\delta_{z_\ell(t)}$.
Define $\bar u_t(z)$ and $\bar g_t(z)$ by averaging $u_\ell^c$ and
$g_\ell^c$ over components at $z$, with weights proportional to
$m_\ell(t)$.  For any $f\in C_c^1(\cZ)$,
\begin{align*}
 \frac{\dd}{\dd t}\int f\dd\rho_t
 &=\sum_\ell m_\ell(t)
       \big[\nabla f(z_\ell(t))\cdot u_\ell^c(t)
                    +g_\ell^c(t)f(z_\ell(t))\big]\\
 &=\int\big[\nabla f\cdot\bar u_t+\bar g_t f\big]\dd\rho_t.
\end{align*}
This is the weak continuity--reaction equation, including when several
components occupy the same state.

\emph{Optimality.}
For a positive edge, stationarity of the unregularized OET objective gives
\[
 c_{ij}+\log(r_i/a_i)+\log(s_j/b_j)=0,
 \qquad
 \sqrt{m_{0,ij}m_{1,ij}}\cos\vartheta_{ij}=\gamma_{ij}.
\]
Writing $J(\gamma)$ for the bracketed objective in
Equation~\eqref{eq:qoet}, it follows that
\begin{align*}
 J(\gamma)
 &=\sum_{ij}\gamma_{ij}
       \big[c_{ij}+\log(r_i/a_i)+\log(s_j/b_j)\big]
       +|\mu|+|\nu|-2\sum_{ij}\gamma_{ij}\\
 &=|\mu|+|\nu|-2\sum_{ij}\gamma_{ij}.
\end{align*}
The sums involving logarithms run over positive edges.  Summing the cone
actions of all transported, decay, and creation components gives exactly
$2\delta^2J(\gamma)=\WFR_\delta^2(\mu,h\push\nu)$ by the
OET representation~\citep{liero2018oet}.  Conditional Jensen inequalities
bound the action of $(\rho,\bar u,\bar g)$ above by this sum.  Its endpoint
constraints bound the same action below by
$\WFR_\delta^2(\mu,h\push\nu)$.  Both bounds are therefore equal, and
the marginal path minimizes the WFR action on $[0,1]$.
\end{proof}

\subsection{Empirical horizontal projection}
\label{app:empirical-projection}

\begin{proposition}[Consistency of empirical horizontal projection]
\label{prop:empirical-projector}
Let $(\rho_B,e_B)$ be weighted empirical populations whose mass, spatial and
residual first moments, matrices $S_B,C_B$, and residual second moment
converge to those of $(\rho,e)$.  Suppose
$\mathcal L_S$ is invertible on $\mathfrak{so}(d_x)$.  Then the fitted
coefficients $(a_B,\Omega_B)$ and the minimum weighted losses converge to
their population counterparts.  In particular, this holds almost surely
for i.i.d.\ mass-weighted samples of a fixed residual field with finite
spatial and residual second moments.
\end{proposition}
\begin{proof}
The empirical normal equations are
\[
 a_B=\bar e_B-\Omega_B\bar x_B,\qquad
 \mathcal L_{S_B}(\Omega_B)=C_B-C_B^\top.
\]
Moment convergence implies convergence of $\mathcal L_{S_B}$ in operator
norm.  Since $\mathcal L_S$ is invertible on the finite-dimensional skew
subspace, $\mathcal L_{S_B}$ is invertible for all sufficiently large $B$,
and continuity of inversion yields
\[
 \Omega_B=\mathcal L_{S_B}^{-1}(C_B-C_B^\top)
 \longrightarrow\mathcal L_S^{-1}(C-C^\top)=\Omega^\star,
 \qquad a_B\longrightarrow a^\star.
\]
If $Q_B=\int\norm{e_B}^2\dd\rho_B$, expanding the quadratic objective
at its minimizer gives
\[
 \min F_B=Q_B-M_B\norm{\bar e_B}^2
             -\operatorname{tr}(\Omega_B^\top C_B).
\]
Every term converges by the assumed moment convergence and the coefficient
convergence just proved.  For i.i.d.\ samples from $\rho/M$ with weight
$M/B$, the strong law supplies these moment limits, and integrability of the
cross-moments follows from Cauchy--Schwarz.  The same conclusion applies to
stratified or importance-weighted sampling whenever its weighted moments
satisfy the stated convergence assumptions.
\end{proof}

\subsection{Proof of Lemma~\ref{lem:horizontal}}
\label{app:horizontal-proof}

\textit{Let $h^\star$ attain Equation~\eqref{eq:qwfr}, and let
$\widetilde\rho_t$ be the bridge marginal of an exact qOET solution from
$\mu_0$ to $h^\star\push\mu_1$, with spatial aggregate velocity
$\widetilde u_x$.  Then}
\[
 \widetilde u_x=P^H_{\widetilde\rho_t}\widetilde u_x
 \quad\text{in }L^2(\widetilde\rho_t)\text{ for almost every }t.
\]
\begin{proof}
By Proposition~\ref{prop:bridge-construction}, the bridge marginal is an
ambient WFR geodesic.  Its action is
$D^2=\qWFR_\delta^2([\mu_0],[\mu_1])$, since the endpoint pose is
globally optimal.  Decompose
\[
 \widetilde u_x=P^H_{\widetilde\rho_t}\widetilde u_x+\xi_t,
 \qquad \xi_t(x)=a_t+\Omega_tx.
\]
On any measurable set of times where the fitted coefficients are bounded,
place $\xi_t$ in the free frame term and subtract it from the relative
spatial velocity.  The total velocity and hence the path are unchanged.
Orthogonality of the projection reduces the action by precisely the
squared norm of this rigid component on that set.

To make this truncation precise, choose the minimum-norm coefficients and let
$E_n=\{t\in[1/n,1-1/n]:\norm{a_t}+\norm{\Omega_t}_{\mathrm F}\le n\}$.
The frame $\mathbf1_{E_n}\xi_t$ is integrable, and the tuple with relative
velocity $\widetilde u_x-\mathbf1_{E_n}\xi_t$ is feasible between
$\mu_0$ and $h^\star\push\mu_1$.  Theorem~\ref{thm:dynamic-static}
and the Pythagorean identity give
\[
 D^2\le D^2-\frac{\lambda_x}{2}
           \int_{E_n}\!\int\norm{\xi_t}^2\dd\widetilde\rho_t\dd t
 \le D^2.
\]
The integral is zero for every $n$.  The sets $E_n$ increase to $(0,1)$
up to a null set, so $\xi_t=0$ in $L^2(\widetilde\rho_t)$ for almost
every $t$.  Thus the original bridge marginal is already horizontal.
\end{proof}

\subsection{Proof of Theorem~\ref{thm:cfm}}
\label{app:cfm-proof}

\textit{Fix $t$ and normalize the mass-weighted bridge measure to a
probability law.  Let $U^c\in L^2$ be the spatial bridge velocity,
$\bar U(z)=\E[U^c\mid Z_t=z]$, and assume $\rho_t$ is regular.  For
every $v\in L^2(\rho_t)$,}
\begin{align*}
 \inf_{\xi\in\mathfrak{se}(d_x)}\E\norm{v(Z_t)-U^c-\xi(Z_t)}^2
 &=\inf_{\xi\in\mathfrak{se}(d_x)}
      \E\norm{v(Z_t)-\bar U(Z_t)-\xi(Z_t)}^2+C_t,\\
 C_t&=\E\norm{U^c-\bar U(Z_t)}^2.
\end{align*}
\textit{Consequently, every population minimizer satisfies
$P^H_{\rho_t}v=P^H_{\rho_t}\bar U$.}
\begin{proof}
Set $\eta=U^c-\bar U(Z_t)$.  By conditional expectation,
$\E[\eta\mid Z_t]=0$.  For a fixed rigid field $\xi$, let
$r(Z_t)=v(Z_t)-\bar U(Z_t)-\xi(Z_t)$.  Since $r$ is a function of
$Z_t$ and all fields have finite second moments,
\[
 \E\langle r(Z_t),\eta\rangle
 =\E\big\langle r(Z_t),\E[\eta\mid Z_t]\big\rangle=0.
\]
Expanding the square now gives
\begin{align*}
 \E\norm{v(Z_t)-U^c-\xi(Z_t)}^2
 &=\E\norm{r(Z_t)-\eta}^2\\
 &=\E\norm{r(Z_t)}^2+\E\norm{\eta}^2.
\end{align*}
The second term is independent of $v$ and $\xi$, so taking the infimum
over the rigid subspace proves the identity.  With $M_t=|\rho_t|$,
Proposition~\ref{prop:projector} identifies the remaining term as
\[
 \inf_\xi\E\norm{v(Z_t)-\bar U(Z_t)-\xi(Z_t)}^2
 =M_t^{-1}\norm{P^H_{\rho_t}(v-\bar U)}_{L^2(\rho_t)}^2.
\]
It is nonnegative and vanishes for $v=\bar U$.  Every minimizer must
therefore satisfy the claimed projected equality.  Applying the same
conditional-expectation calculation to the molecular and growth targets,
without the infimum over rigid fields, identifies their minimizers with
their respective conditional means.
\end{proof}

\subsection{Population recovery}
\label{app:recovery}

\begin{corollary}[Population recovery of quotient geodesics]
\label{cor:recovery}
Assume a globally optimal unregularized qOET solution and regular tissues
along its bridge marginal $\rho_t^\star$.  Let $\E_t$ denote expectation
under the normalized mass-weighted bridge law and let $M_t=|\rho_t^\star|$.
Suppose the population version of Equation~\eqref{eq:qfm-loss},
\begin{align}
 \cL_{\mathrm{pop}}=\int_0^1 M_t\Big[&\lambda_x\inf_{\xi\in\mathfrak{se}(d_x)}
       \E_t\norm{v_x(Z_t,t)-U_x^c-\xi(Z_t)}^2\nonumber\\
       &+\lambda_q\E_t\norm{v_q(Z_t,t)-U_q^c}^2
        +\delta^2\E_t|g(Z_t,t)-G^c|^2\Big]\dd t,
 \label{eq:population-loss}
\end{align}
is minimized exactly over all square-integrable fields.  For fields that
depend on the population, this loss evaluates them at $\rho_t^\star$.
Every minimizer induces the same projected spatial velocity, molecular
velocity, and growth $\rho_t^\star$-almost everywhere for almost every
$t$, and $\rho_t^\star$ is a weak solution of
\begin{equation}
 \partial_t\rho_t+\nabla_x\!\cdot\!\left(\rho_tP_{\rho_t}^H
 v_{\theta,x}\right)+\nabla_q\!\cdot(\rho_tv_{\theta,q})=g_\phi\rho_t.
 \label{eq:population-sampler}
\end{equation}
Writing $D=\qWFR_\delta([\mu_0],[\mu_1])$, this path has action $D^2$ and
\[
 \qWFR_\delta([\rho_s^\star],[\rho_t^\star])=(t-s)D
 \qquad(0\le s<t\le1),
\]
with $\rho_1^\star=h^\star\push\mu_1\in[\mu_1]$.
If Equation~\eqref{eq:population-sampler} has a unique solution with
initial measure $\mu_0$ in the class under consideration, exact population
integration recovers $\rho_t^\star$.
\end{corollary}
\begin{proof}
Let $(\bar u_x,\bar u_q,\bar g)$ be the mass-weighted aggregate bridge
fields.  Theorem~\ref{thm:cfm} and its molecular and growth counterparts
write the population loss as a constant plus
\[
 \int_0^1\!\int\Big[
   \lambda_x\norm{P^H_{\rho_t^\star}(v_x-\bar u_x)}^2
   +\lambda_q\norm{v_q-\bar u_q}^2
   +\delta^2|g-\bar g|^2\Big]\dd\rho_t^\star\dd t.
\]
The aggregate fields attain zero excess loss.  Since the weights
$\lambda_x,\lambda_q,\delta^2$ are positive, every exact minimizer must
make each squared residual zero almost everywhere.  Lemma~\ref{lem:horizontal}
then gives
\[
 P^H_{\rho_t^\star}v_x=\bar u_x,\qquad
 v_q=\bar u_q,\qquad g=\bar g
 \quad\rho_t^\star\text{-almost everywhere, for almost every }t.
\]
Proposition~\ref{prop:bridge-construction} supplies the weak continuity
equation for the aggregate fields.  Substituting these equalities proves
Equation~\eqref{eq:population-sampler} on the same path.

The ambient bridge path is a constant-speed WFR geodesic of length $D$.
For $0\le s<t\le1$, the quotient metric is bounded by the ambient one, so
\begin{align*}
 D&\le\qWFR_\delta([\rho_0^\star],[\rho_s^\star])
     +\qWFR_\delta([\rho_s^\star],[\rho_t^\star])
     +\qWFR_\delta([\rho_t^\star],[\rho_1^\star])\\
  &\le sD+(t-s)D+(1-t)D=D.
\end{align*}
Equality of the sums forces equality in each term.  Its action is $D^2$
by the bridge construction and horizontality.  To obtain the prescribed
terminal representative $\mu_1$, choose a smooth rigid path $H_t$ from
$I$ to $(h^\star)^{-1}$ and set
$\widehat\rho_t=H_t\push\rho_t^\star$.  The construction in
Theorem~\ref{thm:dynamic-static} puts the frame velocity into the free
term, giving a feasible tuple for Equation~\eqref{eq:qaction} with the
same action.  Finally, $\rho_t^\star$ is a solution with initial measure
$\mu_0$, and uniqueness, when assumed, identifies it with the exactly
integrated population solution.
\end{proof}

\subsection{Multiple snapshots}
\label{app:multitime}

\begin{proposition}[Multi-snapshot quotient action]
\label{prop:multitime}
For $t_0<\cdots<t_K$, let $\overline{\mathcal E}$ be the action of
Equation~\eqref{eq:qaction} integrated over physical time.  For nonzero
finite atomic orbit constraints $\rho_{t_k}\in[\mu_k]$,
\begin{equation}
 \inf_{\rho_{t_k}\in[\mu_k]}\overline{\mathcal E}(\rho)
 =\sum_{k=0}^{K-1}
 \frac{\qWFR_\delta^2([\mu_k],[\mu_{k+1}])}{t_{k+1}-t_k}.
 \label{eq:multitime}
\end{equation}
Each summand is attained by a time-rescaled quotient geodesic.  Consecutive
segments admit representatives meeting at their shared snapshot, and,
when the tissues along the segments are regular, their horizontal lifts
concatenate.
\end{proposition}
\begin{proof}
Let $\Delta_k=t_{k+1}-t_k$ and $\tau=(t-t_k)/\Delta_k$.
For a trajectory restricted to this interval, the fields in local time
are $u^\tau=\Delta_k u$ and $g^\tau=\Delta_k g$, with the same scaling
for the frame field.  Changing variables in the action gives
\[
 \mathcal E_{[t_k,t_{k+1}]}
 =\frac{1}{\Delta_k}\mathcal E_{[0,1]}.
\]
Theorem~\ref{thm:dynamic-static} bounds the right-hand side below by
$\qWFR_\delta^2([\mu_k],[\mu_{k+1}])/\Delta_k$.
Summing proves the lower bound in Equation~\eqref{eq:multitime}.

For the upper bound, choose a minimizing quotient geodesic on each unit
interval and rescale it to physical time.  At a shared snapshot, the
incoming endpoint and outgoing initial measure belong to the same orbit.
Push the entire outgoing segment forward by a fixed group element that
makes these two measures coincide.  This leaves its action and terminal
orbit unchanged.  Repeating over the intervals gives a continuous
measure path attaining the sum.  The weak continuity equation holds on
the concatenation because the boundary terms at each shared snapshot
cancel.  For regular horizontal lifts,
Proposition~\ref{prop:frame-covariance} preserves horizontality under
these fixed changes of representative, proving the last assertion.
\end{proof}

\begin{samepage}
\subsection{Architecture covariance}
\label{app:architecture-proof}

\begin{proposition}[Architecture covariance]
\label{prop:architecture}
For $h=(R,b)\in\SE(2)$ and $hz=(Rx+b,q)$, the transformer satisfies
\begin{align*}
 v_{\theta,x}[h_\#\rho](hz,t)&=Rv_{\theta,x}[\rho](z,t),\\
 v_{\theta,q}[h_\#\rho](hz,t)&=v_{\theta,q}[\rho](z,t),\qquad
 g_\phi[h_\#\rho](hz,t)=g_\phi[\rho](z,t).
\end{align*}
Its empirical horizontal field is covariant:
\[
 \big(P^H_{h_\#\rho}v_{\theta,x}[h_\#\rho]\big)(hz)
 =R\big(P^H_\rho v_{\theta,x}[\rho]\big)(z).
\]
Splitting a positive-weight carrier into coincident copies with positive
weights summing to the original weight leaves the attention operator and
the predicted fields unchanged, up to floating-point summation.
\end{proposition}
\end{samepage}
\begin{proof}
\emph{Rigid transformations.}
Let $y_i=x_i-\bar x$.  Under $h$, centered positions and pairwise
differences transform by $R$, while the molecular states, time, and
total mass are unchanged.  Since $R\in SO(2)$,
\[
 \norm{Ry_i}=\norm{y_i},\qquad
 (Ry_i)^\top(Ry_j)=y_i^\top y_j,\qquad
 (Ry_i)^\top J(Ry_j)=y_i^\top Jy_j.
\]
The mass-normalized covariance transforms as
$\Sigma'=R\Sigma R^\top$, so its trace, determinant, and the scalar
$y_i^\top\Sigma y_i$ are also invariant.  Hence the initial embeddings
and pairwise scalar features are invariant.

For one attention head, write the logit before its mass bias as $\ell_{ij}$
and the scalar-channel value as $V_j$.  Its output and normalizer are
\[
 Z_i=\sum_jm_j e^{\ell_{ij}},\qquad
 A_i=\frac{\sum_jm_j e^{\ell_{ij}}V_j}{Z_i},\qquad \log Z_i.
\]
Invariant input features produce invariant queries, keys, values, and
logits, so both $A_i$ and $\log Z_i$ are invariant.  The residual and
pointwise feedforward operations preserve this property.  Induction over
the blocks therefore proves invariance of all hidden scalar features and
of the molecular and growth heads.

Every basis vector of the spatial head transforms by $R$.  For the
centroid-relative and covariance bases this follows from
\[
 y_i'=Ry_i,\qquad Jy_i'=RJy_i,\qquad
 \Sigma'y_i'=R\Sigma y_i,
\]
where $RJ=JR$ for proper planar rotations.  Pairwise differences and
their quarter turns obey the same rule, and the coefficients used to
aggregate them are invariant.  The spatial output is thus covariant.
Proposition~\ref{prop:frame-covariance} then proves covariance of its
horizontal projection.

\emph{Splitting a carrier.}
Replace $(z_j,m_j)$ by copies $(z_j,m_{j,r})$ with
$m_{j,r}>0$ and $\sum_r m_{j,r}=m_j$.  This preserves the tissue measure,
its total mass, centroid, and covariance.  No node input depends on the
individual weight, so all copies have the original node embedding and
identical queries, keys, and values.  For every query $i$,
\[
 \sum_r m_{j,r}e^{\ell_{ij}}=m_je^{\ell_{ij}},\qquad
 \sum_r m_{j,r}e^{\ell_{ij}}V_j=m_je^{\ell_{ij}}V_j.
\]
The split leaves both the numerator and denominator of attention, and
therefore its log-normalizer, unchanged.  Copies used as queries likewise
receive identical updates.  Induction over blocks preserves the equality
of their embeddings.  The output heads depend on these embeddings,
moment-based directions, and attention-weighted pairwise directions, and
each such weighted sum is unchanged by the same identities.  This proves
the splitting assertion.
\end{proof}

\section{Implementation and reproducibility details}
\label{app:implementation}

\subsection{From population guarantees to the implementation}
\label{app:deferred}

The implementation differs from the setting of the population results in
Appendix~\ref{app:proofs} in three ways.
First, the pose search in Equation~\eqref{eq:qoet} is nonconvex, so the
primal--dual gap criterion guarantees optimality of the coupling at the selected pose but not global optimality of the pose itself.
Second, the loss in Equation~\eqref{eq:qfm-loss} fits the rigid field on each
sampled batch, whereas Equation~\eqref{eq:population-loss} fits it to the
population residual.  Proposition~\ref{prop:empirical-projector} gives
convergence for a fixed residual field as batch moments converge, but at
finite batch size the fit can be biased, and because attention is also
computed over the sampled population, the batch size affects the evaluated
field as well as the gradient variance.  Third, generation uses a fixed set
of carriers.  When a characteristic representation of the population
equation exists, its positive-weight carriers satisfy
\begin{equation}
 \dot X_t=P_{\rho_t}^Hv_{\theta,x},\qquad
 \dot Q_t=v_{\theta,q},\qquad
 \frac{\dd}{\dd t}\log w_t=g_\phi.
 \label{eq:sampler}
\end{equation}
Splitting mass can require several carriers at the same initial state with
weights summing to the original mass, and pure creation can require
components with zero initial mass.  A fixed set of carriers therefore
reproduces the population marginal of Corollary~\ref{cor:recovery} only when
it can represent these characteristics, and the implemented RK4 scheme is a
finite-carrier discretization of Equation~\eqref{eq:sampler}.  The analytic
bridge derivatives use local time, and training divides velocity and growth
by the physical interval length.

\begin{algorithm}[htbp]
\caption{Horizontal flow matching with \method{}}
\label{alg:train}
\begin{algorithmic}[1]
\Require measures $\{\widehat\mu_k\}_{k=0}^K$, WFR scale $\delta$,
ground weights $\lambda_x,\lambda_q$
\For{$k=0,\ldots,K-1$}
  \State solve joint $\SE(2)$ qOET (Eq.~\ref{eq:qoet}) for
  $(h_k,\gamma_k)$ and form the source and target masses of Eq.~\ref{eq:semicoupling}
  \State construct transported, decay, and creation bridge components
\EndFor
\For{optimization step $s=1,\ldots,S$}
  \State sample interval $k$, one shared local time $t$, and $B$ cell states from the bridges
  \State evaluate analytic $(z_t,m_t,u_t^c,g_t^c)$
  (Eqs.~\ref{eq:mass-path}--\ref{eq:teacher})
  \State evaluate equivariant $v_\theta$ and invariant $g_\phi$
  \State fit the shared rigid field $(a_B,\Omega_B)$ by weighted least squares
  \State update $(\theta,\phi)$ with Equation~\eqref{eq:qfm-loss}
\EndFor
\State \textbf{infer:} integrate $\dot X=P^H_\rho v_{\theta,x}$,
$\dot Q=v_{\theta,q}$, $\dot w=g_\phi w$, recomputing $P^H_\rho$ at every stage
\end{algorithmic}
\end{algorithm}

\FloatBarrier
\subsection{Numerical and experimental settings}

\begin{table*}[t]
\centering
\caption{\textbf{Shared \method{} configuration.}  One architecture and
optimizer are used on every real dataset (9,000 updates, width 384, five
blocks, eight heads, feed-forward width 768, initial
learning rate $1.5\!\times\!10^{-3}$ with cosine decay to
$2\!\times\!10^{-5}$, EMA 0.995, gradient clip 100, $\delta=10$,
$\lambda_x=1$, and projected RK4 maximum normalized step 0.25).  $B$ is the
number of bridge states per update.  Figures use seed 17, and
Table~\ref{tab:main-results} also includes seeds 29 and 43.  \rev{\textsuperscript{$\ast$}For Drosophila, section depth.}}
\label{tab:hparams}
\resizebox{\textwidth}{!}{%
\begin{tabular}{lclrrrl}
\toprule
Dataset & Times\rev{\textsuperscript{$\ast$}} & Full cells per snapshot & $B$ & $\lambda_q$ & Reference seed & Time input \\
\midrule
Axolotl & 2, 5, 10, 15, 20 & 7,668, 8,106, 9,440, 9,676, 11,319 & 1,024 & 0.15 & 17 & piecewise Fourier \\
Chicken & 4, 7, 10, 14 & 147, 528, 908, 1,967 & 768 & 0.05 & 17 & piecewise Fourier \\
Drosophila & 5.6, 6.3, 7, 9.1, 10.5 & 1,263, 1,248, 1,039, 1,078, 1,110 & 768 & 0.05 & 17 & piecewise Fourier \\
Zebrafish & 5.25, 10, 12, 18, 24 & 563, 1,036, 2,081, 3,048, 5,271 & 1,024 & 0.05 & 17 & piecewise Fourier \\
\bottomrule
\end{tabular}}
\end{table*}

\paragraph{Solving qOET.}
For each interval, a deterministic weighted coreset of at most 1,024 cells
provides initial poses for several optimization starts, after which the coupling solve
uses all cells.  At a fixed pose, neighbours under the ground metric
initialize a sparse edge set, accelerated projected gradients solve the
restricted OET problem, and blockwise scans add every edge with a
reduced-cost violation.  Column generation stops when the relative
primal--dual gap over the full support falls below $10^{-4}$, and cut-locus
masking excludes forbidden transport.  At a fixed coupling, iteratively
reweighted least squares Procrustes~\citep{gower1975generalized,umeyama1991least}
proposes a proper rotation and translation.  Two such updates alternate with
coupling solves, and each is accepted only if the joint objective decreases.
The lowest objective over initial poses is retained, and all 15 intervals
in the real datasets converge within the $10^{-4}$ gap tolerance.  Couplings remain fixed during
neural training.

\paragraph{Network.}
Node inputs are the molecular PCs, the time encoding, the centred radius and
covariance quadratic form of the cell, the log total mass, and the trace and
determinant of the mass-weighted covariance, with no individual weight as a node
input.
Pairwise features are radial distances, centred dot products, and signed
centred areas.  The transformer has five residual blocks with 384 channels,
eight attention heads, QK normalization, SwiGLU, and LayerScale.  As a numerical
check of Proposition~\ref{prop:architecture}, we evaluate each trained
checkpoint in double precision on up to 256 observed cells with unequal
weights and split 64 of them into coincident copies with weight fractions
0.3 and 0.7.  Across the four datasets, the largest change in spatial
velocity, molecular velocity, or growth at any original or copied cell is
below $1.5\times10^{-13}$, whereas the largest spatial velocities are
7.6--42.
Duplicating the same cells without splitting their weight, which changes
the measure, changes the spatial velocity by 0.65--6.2.

\paragraph{Training.}
The optimizer is Adam with cosine decay of the learning rate to
$2\times10^{-5}$.  Training cycles through shuffled intervals, draws local
time by scrambled stratified sampling, and evaluates an exponential moving
average of the weights.  Fitting times for every method are measured on the same single GPU.

\paragraph{Preprocessing.}
Raw counts are library-normalized to the global median, transformed with
$\log(1+x)$, standardized over at most 2,000 prevalence-filtered
high-variance genes, and mapped to a joint 32-dimensional PCA for each
dataset.  Spatial coordinates keep their recorded
orientation and origin and are divided by the median, across snapshots, of
the within-snapshot median 30-neighbour distance.  The zebrafish series
starts at 5.25\,hpf because the retained sections then share a common
embryonic anatomy, whereas the 3.3\,hpf sample contains a distinct
extra-embryonic structure.  stVCR uses its published sampling schedule, and
STORIES 0.1.1 uses its published normalization and hyperparameters on all
four series.

\begin{figure*}[t]
\centering
\includegraphics[width=.88\textwidth]{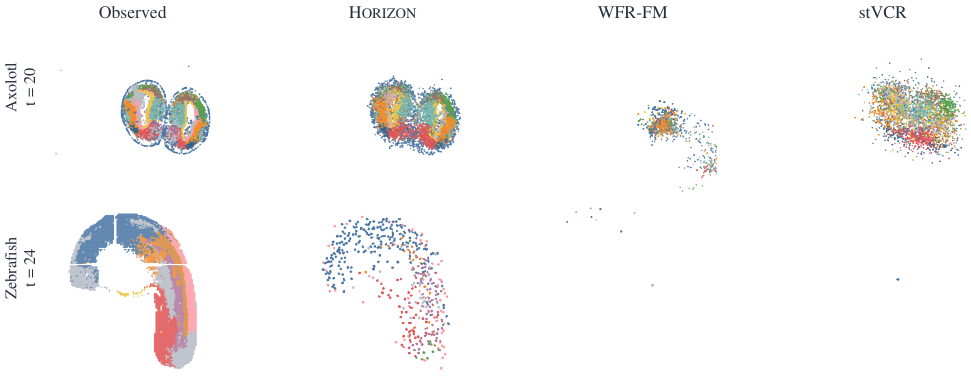}
\caption{\textbf{Spatial comparison at the final snapshot.} All methods use
the same inputs and display canvas, and predictions are integrated from the
first snapshot and aligned by the evaluation pose.  Each displayed point
carries equal mass so point count shows total mass, and colour shows the shared
molecular PCs.}
\label{fig:real-spatial-comparison}
\end{figure*}

\paragraph{Metric definitions.}
SW is weighted sliced $W_2$ with 128 fixed random projections.  MMD is the
square root of biased weighted RBF MMD$^2$ with a pooled median-distance
bandwidth.  Chamfer is the square root of the average of the two
probability-weighted nearest-neighbour squared distances.  Joint metrics use
the same scaled $(x,q)$ ground blocks as the method.  Before every quotient
metric, a scorer independent of qOET searches 72 proper rotation angles,
refines eight candidates, and profiles translation through weighted
centroids using spatial Chamfer.  The same alignment is applied to every
method.

\section{Details of the frame-perturbation benchmark}
\label{app:formal-simulation}

\paragraph{Setup.}
The benchmark starts from the Rectangle trajectory distributed with stVCR.
Three independently sampled intrinsic trajectories use seeds 1103, 2207, and
3301, and frame seeds 401, 503, and 607 generate the mild and severe
proper-$\SE(2)$ perturbations.  The 21 training datasets comprise three
canonical, nine mild, and nine severe views, and each frame seed is paired
with the canonical view of its intrinsic trajectory.  Each dataset contains
six snapshots and retains half of the available cell states, with a minimum
of 128.  A separate 51-frame intrinsic trajectory provides the ground truth
for visualization, and the reported metrics average the five snapshots after
the first.

The qOET solves use an entropic warm start with regularization
$(0.5,0.2,0.08,0.03)$ and 500 Sinkhorn iterations per
stage~\citep{cuturi2013sinkhorn,peyre2019computational} at tolerance
$10^{-8}$, followed by an unregularized APG solve of up to 50,000
iterations with relative gap tolerance $10^{-4}$.  WFR-FM uses 3,000 updates
and stVCR 500 epochs, their published defaults for this trajectory.  The
\emph{no quotient} ablation fixes the pose to the identity, keeps the
coupling solver and transformer unchanged, and trains and integrates without
the projector.  The \emph{no growth} ablation evaluates the trained \method{}
model with its reaction field set to zero.

\paragraph{Results.}
Across the three perturbation regimes and 11 metrics, \method{} achieves lower mean error
than WFR-FM on every comparison except canonical direct joint MMD and outperforms stVCR across all metrics in the severe regime as well as all sliced-Wasserstein, MMD, molecular, and mass metrics in the canonical and mild regimes.  \method{}
and stVCR complete all 21 fits and WFR-FM 20.
Tables~\ref{tab:formal-se2-full-sw}--\ref{tab:formal-se2-full-gene-mass}
report all three regimes, and Appendix~\ref{app:sensitivity} reports the
sensitivity to $\delta$ and $\lambda_q$.

\paragraph{Registration without projection.}
For the \emph{registered, no projector} ablation in
Table~\ref{tab:simulation-ladder}, the interval poses $h_k$ from qOET are
composed to align all snapshots with the first frame, as in the aligned
NicheFlow comparison.  WFR bridges are then built at the identity pose, and
the \method{} architecture is trained with the same optimizer and integrated
with the same step size, both without the projector.  The ablation is
compared with a \method{} control trained under matched seeds on the nine severe
variants with the same architecture and hyperparameters, and the control
matches the main benchmark to within $0.004$ on every spatial metric.
Paired over variants, the full model lowers direct spatial SW by
$0.011\pm0.003$, direct joint SW by $0.006\pm0.002$, and direct Chamfer by
$0.013\pm0.003$, while the difference in quotient spatial SW is
$0.003\pm0.003$.  Evaluation alignment removes the accumulated rigid offset,
so the effect of the projector appears in the direct metrics.

\IfFileExists{tables/simulation_results_full.tex}{%
  \begin{table*}[t]
\centering\scriptsize\setlength{\tabcolsep}{4.2pt}
\caption{Direct and quotient sliced-Wasserstein metrics on the frame-perturbation benchmark (mean$\pm$standard error over successful runs, where $n$ reports successes/attempts and lower is better).}
\label{tab:formal-se2-full-sw}
\resizebox{\textwidth}{!}{%
\begin{tabular}{llcrrrr}\toprule
Method & Regime & $n$ & Direct $W_x$ & Direct $W_{x,g}$ & $W_x^{\mathcal Q}$ & $W_{x,g}^{\mathcal Q}$ \\
\midrule
\textsc{Horizon} (ours) & Canonical & 3/3 & $0.111\!\pm\!0.004$ & $0.079\!\pm\!0.005$ & $0.040\!\pm\!9.84\!\times\!10^{-4}$ & $0.035\!\pm\!9.90\!\times\!10^{-4}$ \\
 & Mild & 9/9 & $0.114\!\pm\!0.004$ & $0.082\!\pm\!0.003$ & $0.039\!\pm\!0.002$ & $0.037\!\pm\!0.002$ \\
 & Severe & 9/9 & $0.115\!\pm\!0.003$ & $0.081\!\pm\!0.002$ & $0.042\!\pm\!0.002$ & $0.037\!\pm\!0.001$ \\
\addlinespace[1.5pt]
WFR-FM & Canonical & 2/3 & $0.223\!\pm\!0.165$ & $0.295\!\pm\!0.222$ & $0.219\!\pm\!0.162$ & $0.293\!\pm\!0.221$ \\
 & Mild & 9/9 & $0.337\!\pm\!0.007$ & $0.227\!\pm\!0.004$ & $0.075\!\pm\!0.003$ & $0.087\!\pm\!0.005$ \\
 & Severe & 9/9 & $1.213\!\pm\!0.027$ & $0.786\!\pm\!0.016$ & $0.110\!\pm\!0.008$ & $0.142\!\pm\!0.016$ \\
\addlinespace[1.5pt]
stVCR & Canonical & 3/3 & $0.115\!\pm\!0.015$ & $0.084\!\pm\!0.009$ & $0.066\!\pm\!0.001$ & $0.062\!\pm\!0.002$ \\
 & Mild & 9/9 & $0.119\!\pm\!0.005$ & $0.087\!\pm\!0.003$ & $0.067\!\pm\!0.001$ & $0.063\!\pm\!0.001$ \\
 & Severe & 9/9 & $0.249\!\pm\!0.021$ & $0.176\!\pm\!0.012$ & $0.119\!\pm\!0.006$ & $0.100\!\pm\!0.005$ \\
\bottomrule\end{tabular}
}
\end{table*}

\begin{table*}[t]
\centering\scriptsize\setlength{\tabcolsep}{4.2pt}
\caption{Direct and quotient Chamfer and joint-MMD metrics on the frame-perturbation benchmark (mean$\pm$standard error over successful runs, where $n$ reports successes/attempts and lower is better).}
\label{tab:formal-se2-full-kernel}
\resizebox{\textwidth}{!}{%
\begin{tabular}{llcrrrr}\toprule
Method & Regime & $n$ & Direct Chamfer & Direct joint MMD & Quotient Chamfer & Quotient joint MMD \\
\midrule
\textsc{Horizon} (ours) & Canonical & 3/3 & $0.116\!\pm\!0.005$ & $0.023\!\pm\!0.002$ & $0.096\!\pm\!1.91\!\times\!10^{-4}$ & $0.006\!\pm\!2.58\!\times\!10^{-4}$ \\
 & Mild & 9/9 & $0.117\!\pm\!0.004$ & $0.024\!\pm\!0.001$ & $0.095\!\pm\!6.91\!\times\!10^{-4}$ & $0.006\!\pm\!5.75\!\times\!10^{-4}$ \\
 & Severe & 9/9 & $0.119\!\pm\!0.004$ & $0.023\!\pm\!9.84\!\times\!10^{-4}$ & $0.096\!\pm\!5.36\!\times\!10^{-4}$ & $0.006\!\pm\!3.90\!\times\!10^{-4}$ \\
\addlinespace[1.5pt]
WFR-FM & Canonical & 2/3 & $0.273\!\pm\!0.164$ & $0.020\!\pm\!1.37\!\times\!10^{-5}$ & $0.272\!\pm\!0.163$ & $0.019\!\pm\!3.48\!\times\!10^{-4}$ \\
 & Mild & 9/9 & $0.227\!\pm\!0.009$ & $0.085\!\pm\!0.002$ & $0.127\!\pm\!0.003$ & $0.021\!\pm\!0.001$ \\
 & Severe & 9/9 & $0.721\!\pm\!0.026$ & $0.311\!\pm\!0.007$ & $0.148\!\pm\!0.009$ & $0.033\!\pm\!0.006$ \\
\addlinespace[1.5pt]
stVCR & Canonical & 3/3 & $0.100\!\pm\!0.002$ & $0.030\!\pm\!0.005$ & $0.094\!\pm\!9.63\!\times\!10^{-4}$ & $0.017\!\pm\!4.39\!\times\!10^{-4}$ \\
 & Mild & 9/9 & $0.101\!\pm\!9.16\!\times\!10^{-4}$ & $0.031\!\pm\!0.002$ & $0.094\!\pm\!0.001$ & $0.018\!\pm\!6.46\!\times\!10^{-4}$ \\
 & Severe & 9/9 & $0.266\!\pm\!0.024$ & $0.055\!\pm\!0.004$ & $0.137\!\pm\!0.004$ & $0.028\!\pm\!0.002$ \\
\bottomrule\end{tabular}
}
\end{table*}

\begin{table*}[t]
\centering\scriptsize\setlength{\tabcolsep}{5.0pt}
\caption{Gene-space and finite-mass metrics on the frame-perturbation benchmark (mean$\pm$standard error over successful runs, where $n$ reports successes/attempts and lower is better).}
\label{tab:formal-se2-full-gene-mass}
\begin{tabular}{llcrrr}\toprule
Method & Regime & $n$ & $W_g$ & Gene MMD & Mass RE \\
\midrule
\textsc{Horizon} (ours) & Canonical & 3/3 & $0.076\!\pm\!0.008$ & $0.011\!\pm\!0.002$ & $2.68\!\times\!10^{-4}\!\pm\!1.06\!\times\!10^{-4}$ \\
 & Mild & 9/9 & $0.084\!\pm\!0.008$ & $0.012\!\pm\!0.002$ & $2.03\!\times\!10^{-4}\!\pm\!4.93\!\times\!10^{-5}$ \\
 & Severe & 9/9 & $0.070\!\pm\!0.005$ & $0.010\!\pm\!9.51\!\times\!10^{-4}$ & $1.92\!\times\!10^{-4}\!\pm\!5.57\!\times\!10^{-5}$ \\
\addlinespace[1.5pt]
WFR-FM & Canonical & 2/3 & $0.484\!\pm\!0.298$ & $0.078\!\pm\!0.007$ & $7.10\!\times\!10^{-4}\!\pm\!9.89\!\times\!10^{-5}$ \\
 & Mild & 9/9 & $0.204\!\pm\!0.011$ & $0.068\!\pm\!0.004$ & $6.23\!\times\!10^{-4}\!\pm\!4.72\!\times\!10^{-5}$ \\
 & Severe & 9/9 & $0.298\!\pm\!0.028$ & $0.115\!\pm\!0.009$ & $0.001\!\pm\!3.00\!\times\!10^{-4}$ \\
\addlinespace[1.5pt]
stVCR & Canonical & 3/3 & $0.127\!\pm\!0.008$ & $0.077\!\pm\!0.002$ & $0.035\!\pm\!0.005$ \\
 & Mild & 9/9 & $0.136\!\pm\!0.004$ & $0.081\!\pm\!0.002$ & $0.035\!\pm\!0.003$ \\
 & Severe & 9/9 & $0.160\!\pm\!0.012$ & $0.132\!\pm\!0.015$ & $0.022\!\pm\!0.002$ \\
\bottomrule\end{tabular}
\end{table*}

%
}{}

\IfFileExists{figures/formal_se2_severe.pdf}{%
\begin{figure*}[t]
\centering
\includegraphics[width=.98\textwidth]{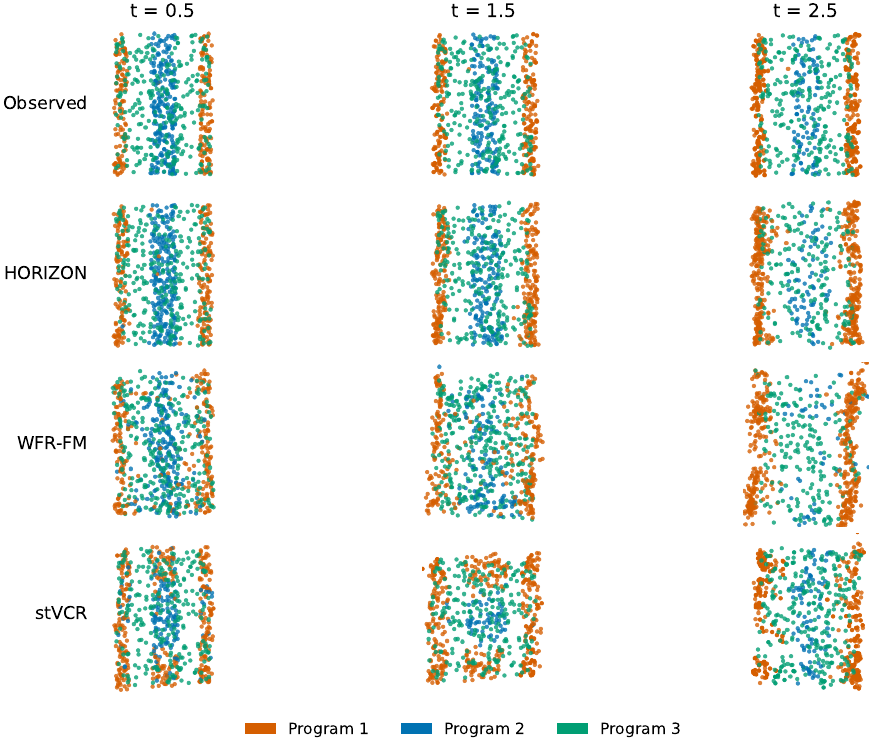}
\caption{\textbf{Trajectories under severe frame perturbations.}
Observed snapshots and predictions of \method{}, WFR-FM, and stVCR at three
times for the first severe variant.  Colors show the three molecular
programs, and all panels share the canvas of the observations after
evaluation alignment.}
\label{fig:formal-se2-severe}
\end{figure*}
}{}

\section{Runtime and baseline comparisons}
\label{app:baseline-details}

All comparisons use the same processed inputs and evaluator.
Table~\ref{tab:main-results} adds the seed standard deviations to the
free-generation block of Table~\ref{tab:real-data}, and
Table~\ref{tab:runtime} gives the fitting times.

\paragraph{Comparison with NicheFlow.}
Even when \method{} generates continuously from the first snapshot while NicheFlow restarts
every interval from its observed source, \method{} achieves lower error than both NicheFlow
variants across all metrics on axolotl and Drosophila and on joint and spatial SW on zebrafish, trailing mainly on chicken where restarts prevent multi-stage error accumulation (Table~\ref{tab:nicheflow}).  Aligning NicheFlow's inputs
with the qOET poses gives worse results than the raw inputs on three of the
four series.

\IfFileExists{tables/main_results.tex}{\begin{table*}[t]
\centering
\caption{\textbf{Free-generation results of Table~\ref{tab:real-data} with
seed variation.}  Lower is better.  \method{} and WFR-FM report mean $\pm$
sample standard deviation over three seeds, and stVCR uses one seed.}
\label{tab:main-results}
\resizebox{\textwidth}{!}{%
\begin{tabular}{llrrrrrrr}
\toprule
Dataset & Method & Joint SW & Joint MMD & Spatial SW & Chamfer & Gene SW & Gene MMD & Mass RE (\%) \\
\midrule
 \multirow{3}{*}{Axolotl} & \method{} & \best{0.0629 $\pm$ 0.0049} & \best{0.0034 $\pm$ 0.0008} & \best{0.2400 $\pm$ 0.0223} & \best{0.3667 $\pm$ 0.0103} & \best{0.1218 $\pm$ 0.0037} & \best{0.0178 $\pm$ 0.0005} & \best{0.0912 $\pm$ 0.0173} \\
  & WFR-FM & 1.4411 $\pm$ 0.1103 & \second{0.1669 $\pm$ 0.0294} & 5.9074 $\pm$ 0.4223 & 2.8603 $\pm$ 0.4860 & 1.1715 $\pm$ 0.2544 & 0.2137 $\pm$ 0.0466 & 146.2522 $\pm$ 67.1410 \\
  & stVCR & \second{1.0405} & 0.2180 & \second{4.2944} & \second{1.0078} & \second{0.2482} & \second{0.0436} & \second{1.2956} \\
\midrule
 \multirow{3}{*}{Chicken} & \method{} & \best{0.1341 $\pm$ 0.0071} & \best{0.0367 $\pm$ 0.0011} & \best{0.1994 $\pm$ 0.0269} & \best{0.3916 $\pm$ 0.0099} & \best{0.5971 $\pm$ 0.0067} & \best{0.0982 $\pm$ 0.0079} & \best{0.4523 $\pm$ 0.1870} \\
  & WFR-FM & \second{0.1873 $\pm$ 0.0343} & \second{0.0637 $\pm$ 0.0178} & \second{0.5397 $\pm$ 0.1989} & \second{0.5761 $\pm$ 0.1281} & \second{0.7513 $\pm$ 0.0141} & \second{0.1352 $\pm$ 0.0049} & \second{1.5590 $\pm$ 0.6506} \\
  & stVCR & 0.5765 & 0.2048 & 1.8059 & 2.0790 & 1.5178 & 0.2617 & 99.1742 \\
\midrule
 \multirow{3}{*}{Drosophila} & \method{} & \best{0.0352 $\pm$ 0.0119} & \best{0.0130 $\pm$ 0.0066} & \best{0.0426 $\pm$ 0.0023} & \best{0.1248 $\pm$ 0.0017} & \best{0.1065 $\pm$ 0.0025} & \best{0.0226 $\pm$ 0.0011} & \best{0.0980 $\pm$ 0.0281} \\
  & WFR-FM & \second{0.0741 $\pm$ 0.0140} & \second{0.0314 $\pm$ 0.0082} & \second{0.0643 $\pm$ 0.0031} & \second{0.1519 $\pm$ 0.0010} & \second{0.1570 $\pm$ 0.0039} & \second{0.0319 $\pm$ 0.0008} & \second{0.2326 $\pm$ 0.0172} \\
  & stVCR & 0.2280 & 0.1119 & 0.8357 & 0.6686 & 0.4273 & 0.1096 & 11.4701 \\
\midrule
 \multirow{3}{*}{Zebrafish} & \method{} & \best{0.0707 $\pm$ 0.0039} & \best{0.0093 $\pm$ 0.0011} & \best{0.2058 $\pm$ 0.0142} & \best{0.3585 $\pm$ 0.0146} & \best{0.2886 $\pm$ 0.0140} & \best{0.0582 $\pm$ 0.0014} & \best{0.5060 $\pm$ 0.0452} \\
  & WFR-FM & \second{1.0349 $\pm$ 0.2718} & \second{0.1603 $\pm$ 0.0285} & \second{4.0486 $\pm$ 1.0621} & \second{3.1423 $\pm$ 1.2365} & \second{1.4411 $\pm$ 0.2543} & \second{0.3743 $\pm$ 0.1079} & \second{44.9785 $\pm$ 8.1872} \\
  & stVCR & 1.8620 & 0.3312 & 7.3944 & 7.8549 & 1.8623 & 0.3766 & 99.4662 \\
\bottomrule
\end{tabular}}
\end{table*}
}{}

\IfFileExists{tables/runtime.tex}{\begin{table}[t]
\centering
\caption{\textbf{Fitting time on one GPU, in seconds.}  HORIZON fit includes
the one-time sparse qOET solve and simulation-free neural regression.
stVCR uses its published training procedure.  Speedup compares complete
HORIZON fitting with stVCR.}
\label{tab:runtime}
\resizebox{\columnwidth}{!}{%
\begin{tabular}{lrrrrr}
\toprule
& \multicolumn{3}{c}{\method{}} & stVCR & stVCR/ours \\
Dataset & qOET & NN & fit & NN & fit speedup \\
\midrule
Axolotl & 794.1 & 298.3 & 1,092.4 & 12,424.4 & 11.4$\times$ \\
Chicken & 465.5 & 281.2 & 746.7 & 1,950.3 & 2.6$\times$ \\
Drosophila & 361.4 & 440.3 & 801.7 & 5,704.7 & 7.1$\times$ \\
Zebrafish & 929.3 & 288.8 & 1,218.2 & 4,752.6 & 3.9$\times$ \\
\bottomrule
\end{tabular}}
\end{table}
}{}

\paragraph{Removing the quotient on real data.}
Table~\ref{tab:quotient-off} repeats the \emph{no quotient} ablation of the
frame-perturbation benchmark on the real time courses.  The WFR coupling is
solved with the pose fixed to the identity by dense APG with the same
continuation schedule, a budget of $30{,}000$ iterations, and a target
relative gap of $10^{-4}$.  The network of Table~\ref{tab:hparams} is trained
on these bridges and integrated at the same step size, both without the
projector.  On chicken, removing the quotient more than doubles spatial SW
and gene SW and quadruples the mass error, whereas on Drosophila, whose
frames are nearly aligned, spatial SW changes only from 0.043 to 0.049.  On
zebrafish (last interval, 18 to 24\,hpf, $3{,}048\times5{,}271$ cells) and
axolotl (first interval), the identity-pose coupling reached neither the
$10^{-4}$ gap tolerance nor a relaxed $10^{-3}$ tolerance within
$150{,}000$ iterations, whereas joint optimization of pose and coupling
converges within $10^{-4}$ on all fifteen intervals.  With the pose fixed,
misaligned pairs require long-range transport, and the coupling problem
becomes ill-conditioned.

\IfFileExists{tables/quotient_off.tex}{\begin{table}[t]
\centering\small
\caption{\textbf{Removing the quotient on real data.}  Lower is better.  \emph{no quotient} fixes the pose to the identity and removes the projector, keeping the architecture, optimizer, step size, and evaluator of \method{}, and both rows generate from the first snapshot.  The \method{} rows repeat Table~\ref{tab:real-data}.}
\label{tab:quotient-off}
\setlength{\tabcolsep}{4pt}
\resizebox{\columnwidth}{!}{%
\begin{tabular}{llrrrrrrr}
\toprule
Dataset & Method & Joint SW & Joint MMD & Spatial SW & Chamfer & Gene SW & Gene MMD & Mass RE (\%) \\
\midrule
 \multirow{2}{*}{Chicken} & \method{} & \best{0.134} & \best{0.037} & \best{0.199} & \best{0.392} & \best{0.597} & \best{0.098} & \best{0.45} \\
  & no quotient & 0.242 & 0.070 & 0.448 & 0.527 & 1.234 & 0.205 & 2.06 \\
\midrule
 \multirow{2}{*}{Drosophila} & \method{} & \best{0.035} & \best{0.013} & \best{0.043} & \best{0.125} & \best{0.107} & 0.023 & \best{0.10} \\
  & no quotient & 0.050 & 0.021 & 0.049 & 0.127 & 0.111 & 0.023 & 0.14 \\
\bottomrule
\end{tabular}}
\end{table}
}{}

\paragraph{NicheFlow configuration on chicken.}
NicheFlow fixes one microenvironment budget from the largest snapshot and
samples every snapshot to that budget without replacement, which the
147-cell chicken D4 slice cannot supply at the published grid.  A finer grid
does not help, because the smallest snapshot saturates at one
microenvironment per cell while the budget set by the largest keeps growing.
We therefore coarsen the grid to the finest feasible spacing,
$dx/dy=0.27/0.36$ against the published $0.15/0.20$ at the same aspect
ratio, which lowers the budget from 432 to 138 microenvironments.  In
addition, at the published radius the two earliest chicken snapshots yield a
modal microenvironment of one cell, so we raise the radius to $0.30$, the
smallest value at which every chicken snapshot reaches the neighbourhood
size that the published radius already yields on Drosophila.  Both changes
are confined to chicken.
Table~\ref{tab:nicheflow} reports both NicheFlow input configurations with
their generation-seed variation and fitting times.

\paragraph{STORIES under both evaluators and both protocols.}
STORIES reports molecular metrics through its own evaluator, whose
normalization differs from the shared evaluator, and its standard protocol
restarts generation from each observed source snapshot.  Table~\ref{tab:real-data} evaluates free
generation from the first snapshot, while Table~\ref{tab:stories-both} reports
both protocols under both evaluators, all with seed 17 and the
configuration of Appendix~\ref{app:implementation}.  Under every combination
\method{} has lower gene SW and MMD than STORIES.  On its own Sinkhorn
objective, STORIES is lower on axolotl with interval restarts and on chicken
and zebrafish under both protocols.

\IfFileExists{tables/stories_both_evaluators.tex}{\begin{table}[t]
\centering\small
\caption{\textbf{STORIES under its own evaluator and under the shared evaluator.}  Lower is better (seed 17 throughout).  STORIES' evaluator uses its normalization and Sinkhorn objective, whereas the shared evaluator is the one used in Table~\ref{tab:real-data}, where the free-generation STORIES row appears.  Bold marks the best entry per column and dataset.}
\label{tab:stories-both}
\setlength{\tabcolsep}{4pt}
\resizebox{\columnwidth}{!}{%
\begin{tabular}{llrrrrr}
\toprule
 & & \multicolumn{3}{c}{STORIES evaluator} & \multicolumn{2}{c}{Shared evaluator} \\
\cmidrule(lr){3-5}\cmidrule(lr){6-7}
Dataset & Method & Sinkhorn$_{\epsilon=.1}$ & Gene SW & Gene MMD & Gene SW & Gene MMD \\
\midrule
 \multirow{3}{*}{Axolotl} & \method{} (first) & 0.2132 & \best{0.1248} & \best{0.0184} & \best{0.1178} & \best{0.0173} \\
  & STORIES (first) & 0.2709 & 1.1093 & 0.2101 & 1.0867 & 0.2071 \\
  & STORIES (restarted) & \best{0.1847} & 0.6373 & 0.1332 & 0.6525 & 0.1376 \\
\midrule
 \multirow{3}{*}{Chicken} & \method{} (first) & 0.4407 & \best{0.6750} & \best{0.1089} & \best{0.6019} & \best{0.0969} \\
  & STORIES (first) & \best{0.2654} & 1.4526 & 0.2705 & 1.4761 & 0.2706 \\
  & STORIES (restarted) & 0.2748 & 1.0530 & 0.2257 & 1.0900 & 0.2289 \\
\midrule
 \multirow{3}{*}{Drosophila} & \method{} (first) & \best{0.4213} & \best{0.1150} & \best{0.0247} & \best{0.1037} & \best{0.0214} \\
  & STORIES (first) & 0.5939 & 0.8224 & 0.2535 & 0.7969 & 0.2407 \\
  & STORIES (restarted) & 0.5696 & 0.5413 & 0.1579 & 0.5609 & 0.1604 \\
\midrule
 \multirow{3}{*}{Zebrafish} & \method{} (first) & 0.1100 & \best{0.3251} & \best{0.0669} & \best{0.2861} & \best{0.0578} \\
  & STORIES (first) & 0.0907 & 0.8332 & 0.2158 & 0.8309 & 0.2155 \\
  & STORIES (restarted) & \best{0.0813} & 0.3833 & 0.1069 & 0.4342 & 0.1219 \\
\bottomrule
\end{tabular}}
\end{table}
}{}

\IfFileExists{tables/nicheflow_results.tex}{\begin{table*}[t]
\centering
\caption{\textbf{NicheFlow with generation-seed variation and fitting time.}
Lower is better, and all rows use the shared evaluator. \method{} generates from
the first snapshot (seed 17), and NicheFlow restarts every
interval from its observed source (Table~\ref{tab:real-data} compares both
methods under this protocol). NicheFlow is trained for 100,000 steps, and
its values are mean $\pm$ sample standard deviation over five generation
seeds. Its aligned input uses the proper $\SE(2)$ qOET poses.}
\label{tab:nicheflow}
\resizebox{\textwidth}{!}{%
\begin{tabular}{lllrrrrrrr}
\toprule
Dataset & Method & Protocol & Joint SW & Joint MMD & Spatial SW & Chamfer & Gene SW & Gene MMD & Fit (h) \\
\midrule
\multirow{3}{*}{Axolotl} & \method{} & first & 0.0605 & 0.0029 & 0.2292 & 0.3616 & 0.1178 & 0.0173 & 0.30 \\
 & NicheFlow raw & restarted & $0.201\!\pm\!0.005$ & $0.025\!\pm\!0.001$ & $0.843\!\pm\!0.023$ & $0.360\!\pm\!0.003$ & $0.279\!\pm\!0.008$ & $0.060\!\pm\!0.001$ & 9.36 \\
 & NicheFlow aligned & restarted & $0.270\!\pm\!0.001$ & $0.038\!\pm\!0.000$ & $1.055\!\pm\!0.007$ & $0.383\!\pm\!0.004$ & $0.345\!\pm\!0.004$ & $0.076\!\pm\!0.001$ & 9.82 \\
\midrule
\multirow{3}{*}{Chicken} & \method{} & first & 0.1328 & 0.0356 & 0.1855 & 0.3898 & 0.6019 & 0.0969 & 0.21 \\
 & NicheFlow raw & restarted & $0.064\!\pm\!0.002$ & $0.025\!\pm\!0.001$ & $0.198\!\pm\!0.005$ & $0.150\!\pm\!0.002$ & $0.217\!\pm\!0.007$ & $0.042\!\pm\!0.001$ & 6.91 \\
 & NicheFlow aligned & restarted & $0.079\!\pm\!0.001$ & $0.035\!\pm\!0.001$ & $0.268\!\pm\!0.004$ & $0.149\!\pm\!0.001$ & $0.231\!\pm\!0.007$ & $0.047\!\pm\!0.001$ & 6.80 \\
\midrule
\multirow{3}{*}{Drosophila} & \method{} & first & 0.0489 & 0.0206 & 0.0447 & 0.1229 & 0.1037 & 0.0214 & 0.22 \\
 & NicheFlow raw & restarted & $0.059\!\pm\!0.011$ & $0.030\!\pm\!0.006$ & $0.158\!\pm\!0.004$ & $0.140\!\pm\!0.001$ & $0.160\!\pm\!0.002$ & $0.045\!\pm\!0.000$ & 6.57 \\
 & NicheFlow aligned & restarted & $0.088\!\pm\!0.012$ & $0.046\!\pm\!0.006$ & $0.185\!\pm\!0.003$ & $0.142\!\pm\!0.001$ & $0.175\!\pm\!0.002$ & $0.051\!\pm\!0.001$ & 6.75 \\
\midrule
\multirow{3}{*}{Zebrafish} & \method{} & first & 0.0700 & 0.0090 & 0.2072 & 0.3682 & 0.2861 & 0.0578 & 0.34 \\
 & NicheFlow raw & restarted & $0.135\!\pm\!0.002$ & $0.029\!\pm\!0.001$ & $0.537\!\pm\!0.009$ & $0.264\!\pm\!0.004$ & $0.243\!\pm\!0.005$ & $0.058\!\pm\!0.001$ & 7.04 \\
 & NicheFlow aligned & restarted & $0.141\!\pm\!0.002$ & $0.029\!\pm\!0.001$ & $0.561\!\pm\!0.008$ & $0.222\!\pm\!0.003$ & $0.251\!\pm\!0.009$ & $0.052\!\pm\!0.001$ & 6.94 \\
\bottomrule
\end{tabular}}
\end{table*}
}{}

\section{Sparse qOET scaling}
\label{app:qoet-scaling}

\begin{table}[t]
\centering
\caption{\textbf{Sparse qOET scaling at a fixed pose.}
Dense and column-generated solves use the same $10^{-4}$ full-gap tolerance.
Memory is peak allocated GPU GiB, and $\Delta J/J$ and marginal error compare the
two converged solutions.}
\label{tab:qoet-scaling}
\resizebox{\columnwidth}{!}{%
\begin{tabular}{lrrrrrr}
\toprule
Support & \multicolumn{2}{c}{Dense} & \multicolumn{2}{c}{Sparse} & Speedup & Agreement \\
 & sec & GiB & sec & GiB & & $\Delta J/J$ / marg. \\
\midrule
$1{,}024^2$ & 10.3 & 0.14 & 11.2 & 0.08 & 0.92$\times$ & $1.1$e--8 / $2.8$e--6 \\
$2{,}048^2$ & 25.7 & 0.55 & 17.8 & 0.16 & 1.44$\times$ & $9.7$e--7 / $1.0$e--4 \\
$4{,}096^2$ & 110.2 & 2.18 & 30.1 & 0.33 & 3.66$\times$ & $1.6$e--5 / $2.0$e--4 \\
$8{,}192^2$ & 649.1 & 8.69 & 62.0 & 0.77 & 10.48$\times$ & $7.7$e--6 / $1.5$e--4 \\
$3{,}048\times5{,}271$ & 290.8 & 2.18 & 173.0 & 1.35 & 1.68$\times$ & $3.7$e--5 / $1.2$e--4 \\
\bottomrule
\end{tabular}}
\end{table}

Table~\ref{tab:qoet-scaling} compares the blockwise column-generation
solver with dense APG on identical problems at a fixed pose.  Agreement is measured
by objective and endpoint marginals because an optimal coupling need not be
unique.  At $8{,}192^2$ support the sparse solver preserves the objective to
$7.7\times10^{-6}$ relative error and both marginals to
$1.5\times10^{-4}$ relative $\ell_1$ error, while using 11.3-fold less peak
memory and 10.5-fold less wall clock.  The largest full zebrafish interval,
$3{,}048\times5{,}271$, is 1.68-fold faster and uses 38\% less peak memory.
All sparse rows satisfy the same $10^{-4}$ full-support primal--dual gap tolerance
as the qOET solutions on real data.

\section{Sensitivity to the WFR scale and the molecular weight}
\label{app:sensitivity}

\IfFileExists{tables/sensitivity.tex}{\begin{table}[t]
\centering\small
\caption{\textbf{Sensitivity to the WFR scale $\delta$ and the molecular weight $\lambda_q$} on the severe regime of the frame-perturbation benchmark (mean $\pm$ standard error over three variants, one per intrinsic seed, with $n$ completed runs).}
\label{tab:sensitivity}
\setlength{\tabcolsep}{4pt}
\resizebox{\columnwidth}{!}{%
\begin{tabular}{lcrrrrr}
\toprule
Configuration & $n$ & Direct sp.\ SW & Quotient sp.\ SW & Quotient joint SW & Gene SW & Mass RE (\%) \\
\midrule
$\delta=10$, $\lambda_q=0.25$ (reference) & 3 & $0.120\!\pm\!0.003$ & $0.044\!\pm\!0.004$ & $0.038\!\pm\!0.004$ & $0.086\!\pm\!0.019$ & $0.03\!\pm\!0.01$ \\
$\delta=5$ & 3 & $0.081\!\pm\!0.010$ & $0.038\!\pm\!0.001$ & $0.034\!\pm\!0.002$ & $0.069\!\pm\!0.005$ & $0.06\!\pm\!0.01$ \\
$\delta=20$ & 3 & $0.126\!\pm\!0.009$ & $0.052\!\pm\!0.004$ & $0.043\!\pm\!0.003$ & $0.066\!\pm\!0.006$ & $0.00\!\pm\!0.00$ \\
$\lambda_q=0.1$ & 3 & $0.107\!\pm\!0.011$ & $0.038\!\pm\!0.001$ & $0.041\!\pm\!0.005$ & $0.107\!\pm\!0.021$ & $0.02\!\pm\!0.01$ \\
$\lambda_q=0.5$ & 3 & $0.123\!\pm\!0.002$ & $0.043\!\pm\!0.002$ & $0.036\!\pm\!0.001$ & $0.072\!\pm\!0.004$ & $0.01\!\pm\!0.00$ \\
\bottomrule
\end{tabular}}
\end{table}
}{}

Table~\ref{tab:sensitivity} varies the WFR scale and the molecular weight on
three severe variants, one per intrinsic seed, with all other settings held
fixed.  We test $\delta\in\{5,10,20\}$ and $\lambda_q\in\{0.1,0.25,0.5\}$,
varying one parameter at a time.  Quotient spatial SW stays within
$0.038$--$0.052$ and joint SW within $0.034$--$0.043$, compared with $0.044$
and $0.038$ at the reference setting.  Mass RE is $0.06\%$ at $\delta=5$ and
rounds to $0.00\%$ at $\delta=20$.  Gene SW varies more across variants than
across settings.

\section{Timepoint holdout evaluation}
\label{app:holdout}

\rev{\paragraph{Out-of-fold preprocessing.}  Each fold refits, on the remaining
snapshots only, every statistic that pools cells across sections, including the
library-size target, the gene prevalence filter and ranking, the gene means and
scales, the PCA, and the spatial scale.  The held-out cells are projected with
this transform and used only as the evaluation target.  Per-cell operations
(library normalization of one cell, $\log(1+x)$) and the population mass, which
is the observed cell count, are applied as before.  Every interval of the
remaining series is then solved from scratch, including the coreset pose
initialization, so neither the coupling nor the pose sees the held-out
snapshot.  WFR-FM and stVCR receive the same fold inputs with their previous budgets.  For Drosophila, stVCR uses the same vectorized spatial-structure loss
as in its main-table run.  \method{} (seeds 17, 29, 43) and WFR-FM (seeds 11, 29, 61) report
three seeds and stVCR seed 17 (Table~\ref{tab:holdout-strict}).}

\begin{table*}[t]
\centering
\caption{\rev{\textbf{Timepoint holdout with out-of-fold preprocessing.}  Lower is better.  In every fold the library-size target, gene filter and ranking, gene standardization, PCA and spatial scale are fitted without the held-out snapshot, and every interval's qOET is solved from scratch.  \method{} averages seeds 17, 29 and 43, WFR-FM averages seeds 11, 29 and 61, and stVCR uses seed 17.  Bold marks the best entry per fold.}}
\label{tab:holdout-strict}
\scriptsize
\setlength{\tabcolsep}{3.2pt}
\begin{tabular}{lllrrrrrrr}
\toprule
Dataset & Held out & Method & Spatial SW & Chamfer & Gene SW & Gene MMD & Joint SW & Joint MMD & Mass RE (\%) \\
\midrule
\multirow{9}{*}{Axolotl} & \multirow{3}{*}{5 DPI} & \method{} & \best{1.136} & \best{0.476} & \best{0.816} & \best{0.220} & \best{0.363} & \best{0.069} & \best{0.2} \\
 &  & WFR-FM & 2.025 & 0.892 & 0.899 & 0.223 & 0.542 & 0.080 & 18.7 \\
 &  & stVCR & 5.390 & 0.750 & 0.918 & 0.222 & 1.251 & 0.300 & 4.4 \\
 & \multirow{3}{*}{10 DPI} & \method{} & \best{1.201} & \best{0.803} & 2.182 & 0.436 & \best{0.574} & \best{0.106} & \best{10.2} \\
 &  & WFR-FM & 3.394 & 3.057 & \best{1.825} & \best{0.381} & 0.934 & 0.121 & 24.4 \\
 &  & stVCR & 5.162 & 1.317 & 2.168 & 0.488 & 1.367 & 0.268 & 17.0 \\
 & \multirow{3}{*}{15 DPI} & \method{} & \best{1.656} & \best{0.514} & \best{2.855} & \best{0.574} & \best{0.846} & \best{0.154} & 4.5 \\
 &  & WFR-FM & 4.973 & 2.150 & 2.998 & 0.632 & 1.425 & 0.221 & 211.2 \\
 &  & stVCR & 3.097 & 0.523 & 2.858 & 0.614 & 1.023 & 0.227 & \best{0.7} \\
\midrule
\multirow{6}{*}{Chicken} & \multirow{3}{*}{D7} & \method{} & 0.374 & 0.366 & 0.368 & 0.077 & 0.129 & 0.074 & \best{15.8} \\
 &  & WFR-FM & \best{0.343} & \best{0.350} & \best{0.258} & \best{0.065} & \best{0.102} & \best{0.061} & 17.7 \\
 &  & stVCR & 2.336 & 2.346 & 1.393 & 0.394 & 0.661 & 0.341 & 99.8 \\
 & \multirow{3}{*}{D10} & \method{} & \best{0.490} & \best{0.489} & 0.588 & 0.161 & \best{0.181} & \best{0.086} & 12.1 \\
 &  & WFR-FM & 0.887 & 0.728 & \best{0.377} & \best{0.080} & 0.226 & 0.119 & \best{6.3} \\
 &  & stVCR & 3.184 & 3.193 & 1.439 & 0.458 & 0.853 & 0.354 & 99.7 \\
\midrule
\multirow{9}{*}{Drosophila} & \multirow{3}{*}{6.3} & \method{} & \best{0.194} & 0.164 & \best{0.181} & \best{0.052} & 0.158 & 0.080 & \best{8.3} \\
 &  & WFR-FM & 0.204 & \best{0.162} & 0.184 & 0.052 & \best{0.158} & \best{0.079} & 8.4 \\
 &  & stVCR & 1.543 & 1.318 & 0.240 & 0.077 & 0.344 & 0.127 & 28.5 \\
 & \multirow{3}{*}{7} & \method{} & \best{0.240} & \best{0.167} & \best{0.254} & \best{0.075} & \best{0.105} & \best{0.055} & \best{15.5} \\
 &  & WFR-FM & 0.253 & 0.171 & 0.292 & 0.085 & 0.124 & 0.067 & 15.6 \\
 &  & stVCR & 0.831 & 0.537 & 0.388 & 0.112 & 0.221 & 0.119 & 33.9 \\
 & \multirow{3}{*}{9.1} & \method{} & \best{0.061} & \best{0.149} & \best{0.430} & 0.159 & \best{0.155} & 0.087 & \best{0.2} \\
 &  & WFR-FM & 0.072 & 0.153 & 0.433 & \best{0.147} & 0.156 & \best{0.084} & 0.5 \\
 &  & stVCR & 1.073 & 0.706 & 0.528 & 0.197 & 0.269 & 0.154 & 26.3 \\
\midrule
\multirow{9}{*}{Zebrafish} & \multirow{3}{*}{10 hpf} & \method{} & \best{0.745} & \best{0.759} & \best{0.597} & \best{0.196} & \best{0.240} & \best{0.066} & \best{45.6} \\
 &  & WFR-FM & 6.624 & 6.357 & 2.270 & 0.386 & 1.702 & 0.374 & 82.7 \\
 &  & stVCR & 5.879 & 6.779 & 2.327 & 0.733 & 1.526 & 0.362 & 99.9 \\
 & \multirow{3}{*}{12 hpf} & \method{} & \best{0.614} & \best{0.511} & \best{0.549} & \best{0.138} & \best{0.214} & \best{0.054} & \best{31.5} \\
 &  & WFR-FM & 1.408 & 1.450 & 1.636 & 0.342 & 0.474 & 0.098 & 42.2 \\
 &  & stVCR & 6.134 & 6.567 & 1.948 & 0.496 & 1.563 & 0.311 & 100.0 \\
 & \multirow{3}{*}{18 hpf} & \method{} & \best{1.911} & \best{1.284} & \best{0.668} & \best{0.202} & \best{0.459} & \best{0.135} & \best{11.2} \\
 &  & WFR-FM & 3.350 & 2.838 & 4.940 & 0.642 & 1.425 & 0.192 & 3597.8 \\
 &  & stVCR & 6.708 & 7.329 & 1.509 & 0.429 & 1.651 & 0.316 & 99.8 \\
\bottomrule
\end{tabular}
\end{table*}

\rev{Refitting the preprocessing within each fold changes \method{}'s spatial
errors by at most 6.4\% at seed 17.  Gene SW is lower or unchanged on every fold because the PCA is now fitted on fewer, and different, snapshots, so molecular values are
comparable within a fold but not with Table~\ref{tab:holdout-per-fold}.
Across the 11 folds \method{} achieves lower spatial SW and Chamfer than WFR-FM on 9 of 11 folds, with the largest margins on axolotl and zebrafish where the sections differ most in frame, and outperforms stVCR across all metrics save gene SW on axolotl 10 DPI.

On the held-out targets, the mass error of \method{} is lower than that of
WFR-FM and of stVCR on 10 of 11 folds each, and its mean over the held-out
times is the lowest of the three methods on axolotl, Drosophila and
zebrafish.  Absolute mass errors nevertheless remain sizable when growth
accelerates sharply across an omitted snapshot, such as on zebrafish
(11.2--45.6\%), because two flanking counts alone do not determine intermediate
growth rates.}

\paragraph{\rev{Full-series preprocessing.}}
When preprocessing is instead fitted once on the full series, each \method{} fold uses 9,000 training
steps and recomputes qOET across
the gap left by the removed snapshot.  WFR-FM uses its published 3,000
updates with seed 11, the same budget as in the main comparison.  Both
methods exclude the held-out snapshot from coupling and training, score only
that target, and stop generation at its time.

Table~\ref{tab:holdout-per-fold} lists every fold under this setting.  \method{} outperforms WFR-FM across all metrics on chicken and zebrafish and on spatial and joint metrics on axolotl, while most exceptions occur on Drosophila, whose frames are
nearly aligned.  WFR-FM fails the zebrafish 18-hpf fold when learned growth
drives log weights beyond the finite range, and stVCR fails the 10-hpf fold
when learned growth eliminates the population.

\section{Population mass over time}
\label{app:mass-trajectories}

\begin{figure*}[!t]
\centering
\includegraphics[width=.96\textwidth]{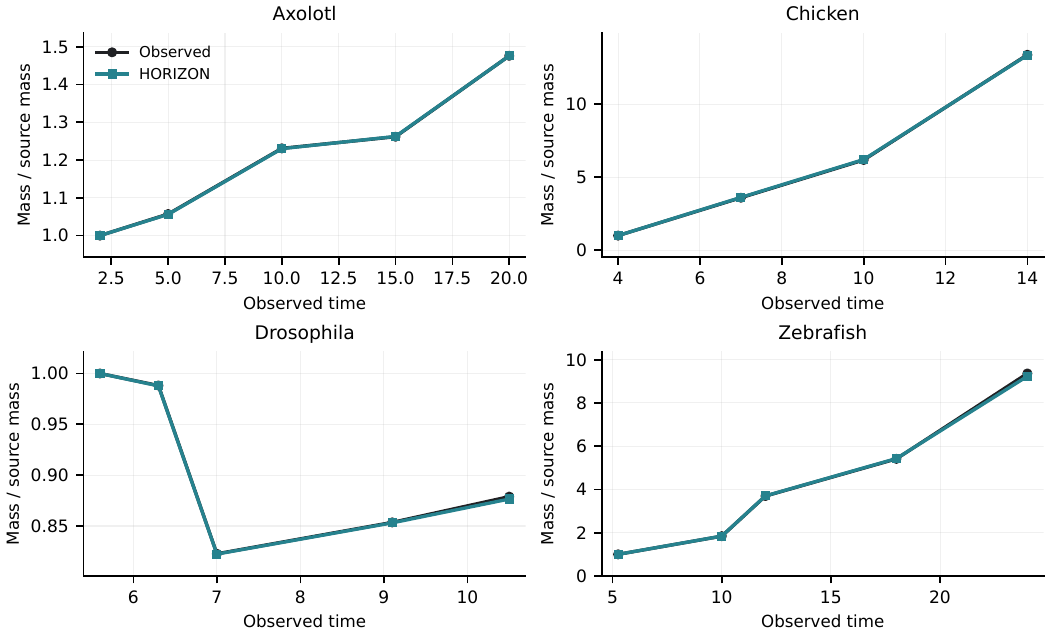}
\caption{\textbf{Population growth across all real time courses.}
Observed cell count ratios and \method{} predictions are normalized
to the first snapshot.  Curves show the predictive mean over three seeds with sample
standard deviation.  The model follows both expansion and contraction, with mean
relative error over target snapshots ranging from 0.091\% to 0.506\%.}
\label{fig:mass}
\end{figure*}

\IfFileExists{tables/holdout_per_fold.tex}{\begin{table*}[t]
\centering
\caption{\textbf{Results for individual holdout folds\rev{ with full-series preprocessing}.} Lower is better. Each row is an independently trained fold.  \rev{Preprocessing was fitted once on the full series, whereas Table~\ref{tab:holdout-strict} refits it within each fold.}}
\label{tab:holdout-per-fold}
\scriptsize
\setlength{\tabcolsep}{3.2pt}
\begin{tabular}{lllrrrrrr}
\toprule
Dataset & Held out & Method & Spatial SW & Chamfer & Gene SW & Gene MMD & Joint SW & Joint MMD \\
\midrule
\multirow{9}{*}{Axolotl} & \multirow{3}{*}{5 DPI} & \method{} & \best{1.1144} & \best{0.4706} & \second{1.1985} & \second{0.2801} & \best{0.4030} & \best{0.07918} \\
 &  & WFR-FM & \second{2.0499} & 0.8889 & \best{1.1741} & 0.2824 & \second{0.5615} & \second{0.09006} \\
 &  & stVCR & 5.2877 & \second{0.7662} & 1.2062 & \best{0.2793} & 1.2498 & 0.2958 \\
 & \multirow{3}{*}{10 DPI} & \method{} & \best{1.1916} & \best{0.7883} & \second{2.2785} & \second{0.4573} & \best{0.6121} & \best{0.1125} \\
 &  & WFR-FM & \second{3.2498} & 2.6579 & \best{1.8388} & \best{0.3835} & \second{0.8631} & \second{0.1261} \\
 &  & stVCR & 5.0112 & \second{1.3247} & 2.3118 & 0.4996 & 1.3296 & 0.2670 \\
 & \multirow{3}{*}{15 DPI} & \method{} & \best{1.6507} & \best{0.5153} & \best{2.8597} & \best{0.6156} & \best{0.7702} & \best{0.1659} \\
 &  & WFR-FM & 5.4997 & 1.4054 & \second{2.9679} & \second{0.6321} & 1.5301 & \second{0.2545} \\
 &  & stVCR & \second{3.6716} & \second{0.6001} & 3.1105 & 0.6536 & \second{1.1758} & 0.2660 \\
\midrule
\multirow{6}{*}{Chicken} & \multirow{3}{*}{D7} & \method{} & \best{0.3626} & \best{0.3540} & \best{0.8776} & \best{0.1885} & \best{0.1726} & \best{0.09744} \\
 &  & WFR-FM & \second{0.3676} & \second{0.3695} & \second{1.0686} & \second{0.2188} & \second{0.1986} & \second{0.1011} \\
 &  & stVCR & 1.7355 & 1.8389 & 1.6706 & 0.2659 & 0.5704 & 0.2562 \\
 & \multirow{3}{*}{D10} & \method{} & \best{0.5036} & \best{0.4958} & \best{0.6841} & \best{0.1625} & \best{0.1894} & \best{0.09546} \\
 &  & WFR-FM & \second{0.9197} & \second{0.8415} & \second{1.0346} & \second{0.2262} & \second{0.2362} & \second{0.1155} \\
 &  & stVCR & 1.8973 & 1.8462 & 1.6085 & 0.2929 & 0.5879 & 0.2338 \\
\midrule
\multirow{9}{*}{Drosophila} & \multirow{3}{*}{6.3} & \method{} & \best{0.1929} & \best{0.1624} & \second{0.2587} & \second{0.07411} & \best{0.1746} & \second{0.08301} \\
 &  & WFR-FM & \second{0.2057} & \second{0.1627} & \best{0.2446} & \best{0.06971} & \second{0.1807} & \best{0.08290} \\
 &  & stVCR & 1.2277 & 1.0658 & 0.3127 & 0.08753 & 0.2863 & 0.1022 \\
 & \multirow{3}{*}{7} & \method{} & \best{0.2458} & \best{0.1668} & \best{0.4997} & \best{0.1240} & \second{0.1597} & \second{0.08124} \\
 &  & WFR-FM & \second{0.2506} & \second{0.1679} & \second{0.5339} & \second{0.1365} & \best{0.1032} & \best{0.05360} \\
 &  & stVCR & 0.8154 & 0.4151 & 0.6412 & 0.1611 & 0.2739 & 0.1488 \\
 & \multirow{3}{*}{9.1} & \method{} & \best{0.06340} & \best{0.1485} & \second{0.6278} & \second{0.2091} & \second{0.2019} & \second{0.1106} \\
 &  & WFR-FM & \second{0.06917} & \second{0.1551} & \best{0.5175} & \best{0.1882} & \best{0.1796} & \best{0.1074} \\
 &  & stVCR & 1.2712 & 0.5366 & 0.6520 & 0.2247 & 0.3649 & 0.2103 \\
\midrule
\multirow{9}{*}{Zebrafish} & \multirow{3}{*}{10 hpf} & \method{} & \best{0.7822} & \best{0.7900} & \best{0.7003} & \best{0.2378} & \best{0.2528} & \best{0.07266} \\
 &  & WFR-FM & \second{5.9320} & \second{5.4335} & \second{1.5381} & \second{0.3999} & \second{1.5076} & \second{0.4154} \\
 &  & stVCR & \multicolumn{6}{c}{failed: learned growth eliminated the population} \\
 & \multirow{3}{*}{12 hpf} & \method{} & \best{0.6474} & \best{0.5200} & \best{0.7653} & \best{0.2055} & \best{0.2367} & \best{0.05822} \\
 &  & WFR-FM & \second{1.7982} & \second{1.6627} & \second{1.1956} & \second{0.2820} & \second{0.5129} & \second{0.1069} \\
 &  & stVCR & 6.2865 & 6.7300 & 1.7878 & 0.3491 & 1.5713 & 0.3018 \\
 & \multirow{3}{*}{18 hpf} & \method{} & \best{1.9884} & \best{1.3160} & \best{0.9462} & \best{0.2680} & \best{0.4896} & \best{0.1417} \\
 &  & WFR-FM & \multicolumn{6}{c}{failed: non-finite growth during generation} \\
 &  & stVCR & \second{5.1215} & \second{5.3531} & \second{1.8569} & \second{0.5967} & \second{1.2778} & \second{0.2929} \\
\bottomrule
\end{tabular}
\end{table*}
}{}

Figure~\ref{fig:mass} shows how the learned reaction field tracks total
mass.  Chicken and zebrafish grow by factors of 13.38 and 9.36, axolotl
reaches 1.48, and \rev{the Drosophila cell count changes little across sections}.

\section{Additional biological analyses}
\label{app:biology}

\begin{figure*}[!t]
\centering
\includegraphics[width=.98\textwidth]{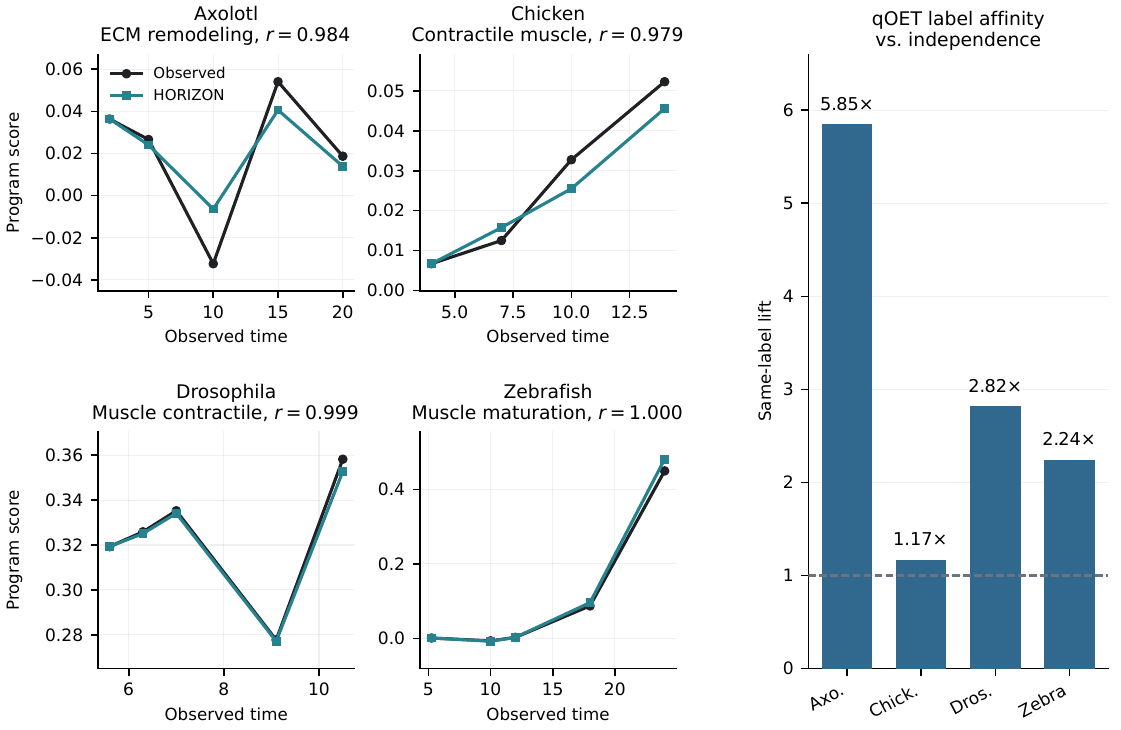}
\caption{\textbf{Biological program dynamics and coupling structure.}  \rev{For Drosophila the horizontal axis is section depth.}
The four panels on the left compare observed and predicted program trajectories.
The panel on the right summarizes label affinity weighted by
transported mass, averaged over intervals and relative to independent marginals.}
\label{fig:biology}
\end{figure*}

To assess whether the couplings preserve biological annotations, we compute
label affinity weighted by transported mass,
$\sum_{ij}\gamma_{ij}\mathbf1[y_i=y_j]/\sum_{ij}\gamma_{ij}$, and divide it by
the probability of matching labels under independent transported marginals.
Mean lift is 5.85 for fine labels in axolotl, 2.24 for fine labels in
zebrafish, 2.82 for tissue labels in Drosophila, and 1.17 for region labels
in chicken.  Lift for
axolotl injury side ranges from 1.92 to 1.97.

The program analyses complement these coupling summaries.  Chicken,
Drosophila, and zebrafish panels use global weighted means of matched
high-variance genes along the generated trajectory.  Chicken programs use HAND1/2,
MYH1F/MYH1B, collagen/PCOLCE, and HBZ.  Drosophila programs use Mhc/actins and
Hsp83.  Zebrafish programs use FOXA, TBX/mesp, myosin light chain, NANOG,
keratin, and HOX sets.  For axolotl, identities of injured and uninjured
cohorts at 2 DPI are propagated by the Lagrangian flow.

Contrasts in
reactive glia (VIM/\allowbreak GFAP/\allowbreak SLC1A3/\allowbreak SOX2/\allowbreak JUN), neuronal maturation
(DCX/\allowbreak MAP2/\allowbreak GAD1/\allowbreak GAD2/\allowbreak SYP), and collagen remodeling are compared with observed
contrasts between injury sides.  The qOET lift for injury labels supports this
cohort interpretation.  Correlations use interval summaries after the initial
snapshot, with $n=4,3,4,4$ for axolotl, chicken, Drosophila, and zebrafish.
Across these program
sets, correlations span 0.948--0.989, 0.979--1.000, 0.989--0.999, and
0.993--1.000, respectively.

\section{Additional spatial trajectories}
\label{app:visuals}

\begin{figure}[!ht]
\centering
{\small\textbf{(a) Chicken heart}}\par\vspace{2pt}
\includegraphics[width=.96\textwidth]{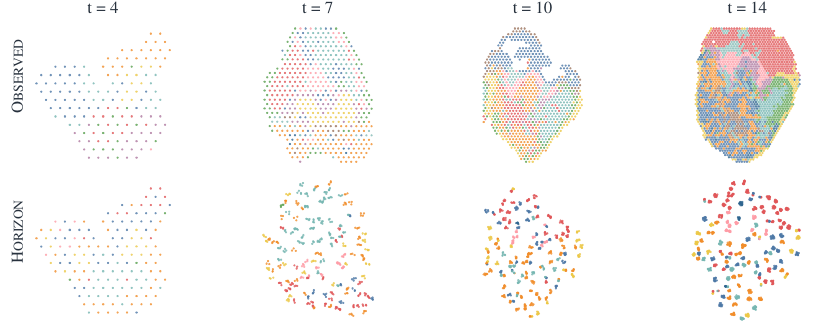}\par\vspace{3pt}
{\small\textbf{(b) Drosophila \rev{embryo, serial sections}}}\par\vspace{2pt}
\includegraphics[width=\textwidth]{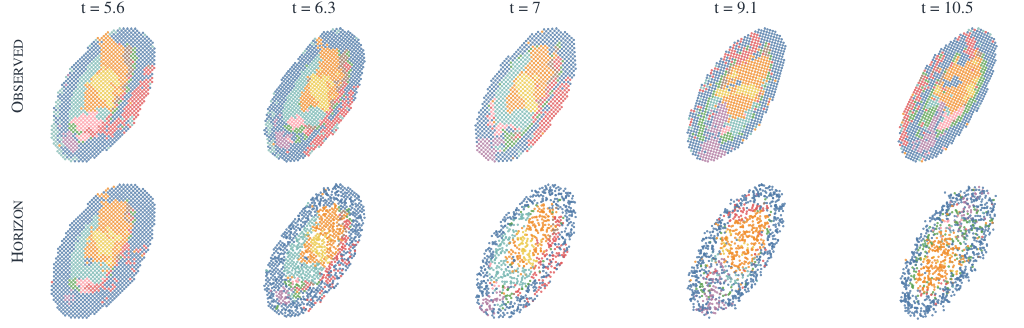}\par\vspace{3pt}
\caption{\textbf{Additional generated time courses.} Zebrafish is shown in Figure~\ref{fig:zebrafish-rollout}.  Observations are on top and
\method{} predictions on the bottom.  Each displayed point carries equal
mass, so point count shows abundance.  Colors show observed annotations on
top and annotations decoded from the predicted molecular PCs on the bottom.}
\label{fig:additional-grids}
\end{figure}

\section{\rev{Registration and projection on real data}}
\label{app:projector-real}

\rev{The four variants of Table~\ref{tab:projector-real} share the precomputed
full-cell qOET bridges, the network of Table~\ref{tab:hparams}, the
optimizer, the step budget, the RK4 step size and the evaluator, and each is
trained with seeds 17, 29 and 43.  For the ambient loss, the bridges of
interval $k$ are mapped to the frame of the first snapshot by the composition
$G_k=h_0\cdots h_{k-1}$ of the qOET poses, the registration also used for the
aligned NicheFlow input.  Because the network is equivariant and the ambient
loss does not change under a common rigid motion of a batch, the frame in
which the bridges are written does not affect what the ambient loss learns, and the comparison therefore isolates the projector.
Table~\ref{tab:projector-real-full} lists all metrics.}

\begin{table*}[t]
\centering
\caption{\rev{\textbf{Registration and projection on real data, all metrics.} Mean $\pm$ standard deviation over seeds 17, 29 and 43.  Quotient metrics (first five columns) use the shared evaluator.  Direct metrics compare with targets registered to the first frame by the composed qOET poses.  \emph{Rotation} is the accumulated angle of the rigid part of the integrated velocity at the last snapshot, and \emph{rigid share} is the fraction of the network's raw spatial kinetic energy carried by a shared rigid motion, averaged along the generated trajectory.  Rows with projection in generation integrate no rigid motion, since the projector removes it.}}
\label{tab:projector-real-full}
\scriptsize
\setlength{\tabcolsep}{2.6pt}
\resizebox{\textwidth}{!}{%
\begin{tabular}{llrrrrrrrrr}
\toprule
Dataset & Variant & Joint SW & Spatial SW & Chamfer & Gene SW & Mass RE (\%) & Direct sp.\ SW & Direct Chamfer & Rotation ($^\circ$) & Rigid share (\%) \\
\midrule
\multirow{4}{*}{Axolotl} & \method{} & 0.063$\pm$0.002 & 0.243$\pm$0.010 & 0.377$\pm$0.001 & 0.123$\pm$0.005 & 0.10$\pm$0.02 & 0.239$\pm$0.001 & 0.378$\pm$0.000 & 0.00$\pm$0.00 & 12.3$\pm$1.3 \\
 & registered, no projector & 0.061$\pm$0.006 & 0.234$\pm$0.025 & 0.365$\pm$0.020 & 0.126$\pm$0.003 & 0.12$\pm$0.02 & 0.228$\pm$0.019 & 0.366$\pm$0.020 & 0.64$\pm$0.02 & 0.2$\pm$0.0 \\
 & projection in loss only & 0.063$\pm$0.002 & 0.243$\pm$0.010 & 0.377$\pm$0.001 & 0.123$\pm$0.005 & 0.10$\pm$0.02 & 0.549$\pm$0.041 & 0.398$\pm$0.004 & 0.82$\pm$0.35 & 12.3$\pm$1.3 \\
 & projection in generation only & 0.061$\pm$0.006 & 0.234$\pm$0.025 & 0.365$\pm$0.020 & 0.126$\pm$0.003 & 0.12$\pm$0.02 & 0.231$\pm$0.020 & 0.366$\pm$0.020 & 0.00$\pm$0.00 & 0.2$\pm$0.0 \\
\midrule
\multirow{4}{*}{Chicken} & \method{} & 0.128$\pm$0.007 & 0.212$\pm$0.006 & 0.401$\pm$0.003 & 0.581$\pm$0.036 & 0.45$\pm$0.07 & 0.215$\pm$0.007 & 0.403$\pm$0.001 & 0.00$\pm$0.00 & 2.3$\pm$1.5 \\
 & registered, no projector & 0.138$\pm$0.006 & 0.211$\pm$0.020 & 0.401$\pm$0.011 & 0.641$\pm$0.011 & 0.45$\pm$0.19 & 0.260$\pm$0.039 & 0.405$\pm$0.012 & 0.94$\pm$1.08 & 0.9$\pm$0.5 \\
 & projection in loss only & 0.129$\pm$0.006 & 0.218$\pm$0.017 & 0.401$\pm$0.003 & 0.581$\pm$0.036 & 0.45$\pm$0.07 & 0.243$\pm$0.022 & 0.407$\pm$0.004 & 2.72$\pm$3.39 & 2.3$\pm$1.5 \\
 & projection in generation only & 0.138$\pm$0.006 & 0.211$\pm$0.020 & 0.401$\pm$0.011 & 0.641$\pm$0.011 & 0.45$\pm$0.19 & 0.213$\pm$0.022 & 0.404$\pm$0.011 & 0.00$\pm$0.00 & 0.9$\pm$0.5 \\
\midrule
\multirow{4}{*}{Drosophila} & \method{} & 0.034$\pm$0.013 & 0.043$\pm$0.004 & 0.125$\pm$0.001 & 0.103$\pm$0.002 & 0.06$\pm$0.01 & 0.041$\pm$0.002 & 0.126$\pm$0.001 & 0.00$\pm$0.00 & 36.5$\pm$2.3 \\
 & registered, no projector & 0.035$\pm$0.012 & 0.043$\pm$0.003 & 0.126$\pm$0.001 & 0.107$\pm$0.001 & 0.06$\pm$0.01 & 0.043$\pm$0.002 & 0.127$\pm$0.001 & 0.22$\pm$0.14 & 0.1$\pm$0.0 \\
 & projection in loss only & 0.034$\pm$0.013 & 0.043$\pm$0.004 & 0.125$\pm$0.001 & 0.103$\pm$0.002 & 0.06$\pm$0.01 & 0.281$\pm$0.019 & 0.207$\pm$0.013 & 16.54$\pm$1.39 & 36.5$\pm$2.3 \\
 & projection in generation only & 0.035$\pm$0.012 & 0.043$\pm$0.003 & 0.126$\pm$0.001 & 0.107$\pm$0.001 & 0.06$\pm$0.01 & 0.041$\pm$0.003 & 0.127$\pm$0.001 & 0.00$\pm$0.00 & 0.1$\pm$0.0 \\
\midrule
\multirow{4}{*}{Zebrafish} & \method{} & 0.069$\pm$0.002 & 0.192$\pm$0.012 & 0.348$\pm$0.011 & 0.283$\pm$0.027 & 0.46$\pm$0.17 & 0.192$\pm$0.012 & 0.349$\pm$0.010 & 0.00$\pm$0.00 & 9.0$\pm$1.4 \\
 & registered, no projector & 0.071$\pm$0.002 & 0.205$\pm$0.010 & 0.349$\pm$0.003 & 0.289$\pm$0.007 & 0.50$\pm$0.03 & 0.208$\pm$0.013 & 0.351$\pm$0.004 & 0.50$\pm$0.17 & 0.1$\pm$0.0 \\
 & projection in loss only & 0.069$\pm$0.002 & 0.192$\pm$0.012 & 0.348$\pm$0.011 & 0.283$\pm$0.027 & 0.46$\pm$0.17 & 0.741$\pm$0.176 & 0.512$\pm$0.060 & 6.70$\pm$5.65 & 9.0$\pm$1.4 \\
 & projection in generation only & 0.071$\pm$0.002 & 0.205$\pm$0.010 & 0.349$\pm$0.003 & 0.289$\pm$0.007 & 0.50$\pm$0.03 & 0.201$\pm$0.012 & 0.350$\pm$0.004 & 0.00$\pm$0.00 & 0.1$\pm$0.0 \\
\bottomrule
\end{tabular}}
\end{table*}

\begin{table}[t]
\centering\small
\caption{\rev{\textbf{Properties of the fitted qOET couplings.}  For each interval we report creation and decay mass as a fraction of target mass, the mean number of targets reached from one transported source cell, the fraction of target mass reached from sources that split, and the share of the bridges' spatial kinetic energy carried by a shared rigid motion under the qOET pose and (where converged) under the identity pose.}}
\label{tab:teacher-stats}
\setlength{\tabcolsep}{3pt}
\resizebox{\columnwidth}{!}{%
\begin{tabular}{llrrrrr}
\toprule
Dataset & Interval & Creation+decay (\%) & Targets/source & From split sources (\%) & Rigid share, qOET (\%) & Rigid share, identity (\%) \\
\midrule
\multirow{4}{*}{Axolotl} & 2$\to$5 & 0.000 & 2.13 & 91.6 & 0.030 & -- \\
 & 5$\to$10 & 0.021 & 2.28 & 95.6 & 0.025 & -- \\
 & 10$\to$15 & 0.021 & 2.14 & 83.8 & 0.069 & -- \\
 & 15$\to$20 & 0.018 & 2.28 & 90.9 & 0.420 & -- \\
\midrule
\multirow{3}{*}{Chicken} & 4$\to$7 & 0.000 & 5.80 & 100.0 & 0.005 & 98.09 \\
 & 7$\to$10 & 0.000 & 2.97 & 100.0 & 0.025 & 97.48 \\
 & 10$\to$14 & 0.000 & 3.66 & 100.0 & 0.005 & 77.06 \\
\midrule
\multirow{4}{*}{Drosophila} & 5.6$\to$6.3 & 0.000 & 2.29 & 73.2 & 0.006 & 0.33 \\
 & 6.3$\to$7 & 0.000 & 2.19 & 69.5 & 0.000 & 0.06 \\
 & 7$\to$9.1 & 0.000 & 2.35 & 99.7 & 0.001 & 1.13 \\
 & 9.1$\to$10.5 & 0.000 & 2.44 & 100.0 & 0.004 & 0.06 \\
\midrule
\multirow{4}{*}{Zebrafish} & 5.25$\to$10 & 0.000 & 3.25 & 100.0 & 0.000 & -- \\
 & 10$\to$12 & 0.000 & 3.79 & 100.0 & 0.078 & -- \\
 & 12$\to$18 & 0.000 & 2.83 & 100.0 & 0.065 & -- \\
 & 18$\to$24 & 0.000 & 3.14 & 99.9 & 0.024 & -- \\
\bottomrule
\end{tabular}}
\end{table}

\rev{All four variants agree on quotient shape
metrics within seed variation, with the ambient model slightly lower on axolotl spatial SW and Chamfer alongside a larger seed spread than \method{}.  On direct spatial SW, \method{} improves over the registered, unprojected
model on chicken, Drosophila and zebrafish, and on axolotl and chicken its
direct error varies less across seeds.  The horizontal loss also lowers the
mean gene SW on all four series, most clearly on chicken,
and during training it reaches lower molecular and growth regression
losses on three of them.  This is consistent with the ambient loss spending
capacity on the rigid component of each batch residual, which carries no
information about the tissue.  Finally, because the horizontal loss does not constrain
the rigid part of the field, integrating that part without projection rotates the tissue by
$0.8^\circ$ (axolotl), $2.7^\circ$ (chicken), $16.5^\circ$ (Drosophila) and
$6.7^\circ$ (zebrafish) by the last snapshot.  The ambient model learns little
rigid motion here because the qOET bridges contain almost none
(Table~\ref{tab:teacher-stats}), and its generated trajectory rotates by $0.2$--$0.9^\circ$.  The projector removes such drift.}

\rev{Direct metrics use targets registered by $G_k$ and so inherit any error
of the qOET poses.  The Drosophila embryo is nearly symmetric under a half
turn, and its Chamfer-based evaluation pose can flip by $180^\circ$ at the last
snapshot, which inflates joint metrics in some runs.  Spatial and molecular metrics are unaffected.}

\section{\rev{Initial carrier support}}
\label{app:support}

\rev{Table~\ref{tab:teacher-stats} summarizes the fitted qOET couplings.  Creation
and decay carry at most 0.021\% of the target mass.  At $\delta=10$ almost every
pair lies inside the cut locus, so the unbalanced coupling realizes growth
almost entirely by mass change along transported pairs.  Splitting, by contrast, is common.  One transported source cell
reaches 2.1--5.8 targets, and 70--100\% of the target mass comes from sources
that split.  A deterministic flow cannot split a carrier, and duplicating a
carrier at the same state changes neither the measure nor, by
Proposition~\ref{prop:architecture}, the predicted fields.  More support
therefore requires a smoother initial measure.  We replace each initial cell
by $K$ carriers that keep its molecular state and share its weight, at
positions perturbed by isotropic Gaussian noise whose standard deviation is
$\sigma$ times the median nearest-neighbour spacing of the first snapshot.
This assumes that the first section samples a tissue that is locally
homogeneous at that scale.}

\begin{table}[t]
\centering\small
\caption{\rev{\textbf{Initial carrier support.}  Each initial cell is replaced by $K$ carriers with its molecular state and weight$/K$, at positions jittered by $\sigma$ times the median nearest-neighbour spacing, keeping the trained model and evaluator unchanged.  Values average three jitter seeds, where $K=1$ is the single-carrier baseline.  Drosophila joint SW is omitted because the Chamfer-based evaluation pose of its nearly half-turn-symmetric embryo flips between runs at 10.5.}}
\label{tab:support}
\setlength{\tabcolsep}{3pt}
\resizebox{\columnwidth}{!}{%
\begin{tabular}{lrrrrrrrrr}
\toprule
Dataset & $K$ & $\sigma$ & Carriers & Spatial SW & Chamfer & Gene SW & Gene MMD & Joint SW & Mass RE (\%) \\
\midrule
\multirow{13}{*}{Chicken} & 1 & 0 & 147 & 0.184 & 0.389 & 0.613 & 0.098 & 0.135 & 0.27 \\
 & 2 & 0.25 & 294 & 0.165 & 0.316 & 0.592 & 0.096 & 0.124 & 0.26 \\
 & 2 & 0.5 & 294 & 0.189 & 0.332 & 0.590 & 0.099 & 0.125 & 0.23 \\
 & 2 & 1 & 294 & 0.273 & 0.406 & 0.619 & 0.104 & 0.143 & 0.19 \\
 & 4 & 0.25 & 588 & 0.163 & 0.267 & 0.564 & 0.092 & 0.117 & 0.27 \\
 & 4 & 0.5 & 588 & 0.165 & 0.273 & 0.556 & 0.090 & 0.115 & 0.23 \\
 & 4 & 1 & 588 & 0.295 & 0.375 & 0.608 & 0.096 & 0.141 & 0.20 \\
 & 8 & 0.25 & 1,176 & 0.150 & 0.222 & 0.573 & 0.088 & 0.117 & 0.26 \\
 & 8 & 0.5 & 1,176 & 0.164 & 0.235 & 0.561 & 0.086 & 0.115 & 0.25 \\
 & 8 & 1 & 1,176 & 0.267 & 0.339 & 0.592 & 0.090 & 0.129 & 0.22 \\
 & 16 & 0.25 & 2,352 & 0.148 & 0.196 & 0.562 & 0.081 & 0.114 & 0.27 \\
 & 16 & 0.5 & 2,352 & 0.155 & 0.210 & 0.548 & 0.080 & 0.110 & 0.25 \\
 & 16 & 1 & 2,352 & 0.256 & 0.320 & 0.573 & 0.083 & 0.122 & 0.23 \\
\midrule
\multirow{13}{*}{Zebrafish} & 1 & 0 & 563 & 0.207 & 0.368 & 0.286 & 0.058 & 0.070 & 0.52 \\
 & 2 & 0.25 & 1,126 & 0.191 & 0.315 & 0.286 & 0.058 & 0.066 & 0.46 \\
 & 2 & 0.5 & 1,126 & 0.186 & 0.302 & 0.284 & 0.059 & 0.065 & 0.46 \\
 & 2 & 1 & 1,126 & 0.220 & 0.332 & 0.299 & 0.062 & 0.073 & 0.62 \\
 & 4 & 0.25 & 2,252 & 0.190 & 0.287 & 0.284 & 0.057 & 0.066 & 0.54 \\
 & 4 & 0.5 & 2,252 & 0.188 & 0.271 & 0.283 & 0.057 & 0.066 & 0.64 \\
 & 4 & 1 & 2,252 & 0.213 & 0.303 & 0.291 & 0.059 & 0.071 & 0.74 \\
 & 8 & 0.25 & 4,504 & 0.190 & 0.260 & 0.282 & 0.055 & 0.066 & 0.52 \\
 & 8 & 0.5 & 4,504 & 0.182 & 0.243 & 0.282 & 0.056 & 0.064 & 0.59 \\
 & 8 & 1 & 4,504 & 0.215 & 0.278 & 0.289 & 0.057 & 0.071 & 0.72 \\
 & 16 & 0.25 & 9,008 & 0.187 & 0.242 & 0.283 & 0.054 & 0.065 & 0.52 \\
 & 16 & 0.5 & 9,008 & 0.178 & 0.226 & 0.281 & 0.054 & 0.064 & 0.58 \\
 & 16 & 1 & 9,008 & 0.212 & 0.266 & 0.287 & 0.055 & 0.070 & 0.76 \\
\midrule
\multirow{10}{*}{Drosophila} & 1 & 0 & 1,263 & 0.043 & 0.125 & 0.107 & 0.022 & -- & 0.09 \\
 & 2 & 0.25 & 2,526 & 0.042 & 0.123 & 0.106 & 0.022 & -- & 0.09 \\
 & 2 & 0.5 & 2,526 & 0.047 & 0.128 & 0.105 & 0.022 & -- & 0.09 \\
 & 2 & 1 & 2,526 & 0.057 & 0.138 & 0.106 & 0.023 & -- & 0.09 \\
 & 4 & 0.25 & 5,052 & 0.040 & 0.111 & 0.105 & 0.022 & -- & 0.09 \\
 & 4 & 0.5 & 5,052 & 0.043 & 0.114 & 0.104 & 0.022 & -- & 0.09 \\
 & 4 & 1 & 5,052 & 0.055 & 0.123 & 0.104 & 0.022 & -- & 0.09 \\
 & 8 & 0.25 & 10,104 & 0.041 & 0.104 & 0.105 & 0.022 & -- & 0.09 \\
 & 8 & 0.5 & 10,104 & 0.040 & 0.106 & 0.104 & 0.022 & -- & 0.09 \\
 & 8 & 1 & 10,104 & 0.049 & 0.115 & 0.103 & 0.022 & -- & 0.09 \\
\bottomrule
\end{tabular}}
\end{table}

\rev{Table~\ref{tab:support} uses the trained models and shared evaluator.  On
chicken (147 cells growing to 1,967) and zebrafish (563 to 5,271), $K\ge4$
with $\sigma\le0.5$ lowers Chamfer distance by 22--50\% and also lowers
spatial SW, joint SW and gene SW, while mass error changes by at most
0.12 percentage points.  A larger jitter, $\sigma=1$, blurs the tissue
boundary and raises spatial SW.  On Drosophila, whose cell count decreases,
Chamfer distance drops by up to 17\% and the other errors change little.  All other tables use $K=1$.}

\section{\rev{Initialization of the pose search}}
\label{app:qoet-init}

\rev{For every interval we ran the coreset stage of the pose search
(Section~\ref{sec:method}) from 120 initial poses, comprising 72 rotations at $5^\circ$
spacing with centroid-matched translation and 24 rotations with each of two
random translation offsets of 0.3 tissue radii.  Converged poses within
$20^\circ$ of each other form one orientation.  The best start of the best
orientation and the best start of any other orientation were refined on the
full support with the sparse solver of Appendix~\ref{app:implementation}
and compared with the reference coupling from the 8-start search (Table~\ref{tab:qoet-init}).  The best
solution among the 120 starts serves as the 120-start optimum below.  The search in the main experiments uses eight starting angles.
The table also reports whether it still reaches this orientation when all
eight angles are rotated together by $5^\circ$, $10^\circ$, and so on up to
$40^\circ$.}

\begin{table}[t]
\centering\small
\caption{\rev{\textbf{Initialization of the pose search.}  Per interval, 120 coreset starts (72 rotations at $5^\circ$ spacing with centroid-matched translation and 24 rotations with each of two translation offsets of 0.3 tissue radii).  \emph{Width} is the angular spread of starts within 0.1\% of the best coreset objective, \emph{Other} is the relative objective excess and angle of the best start more than $20^\circ$ away after full-support refinement, and \emph{Replicas} counts how many of the nine rotated versions of the 8-start schedule reach the best orientation.  The last three columns compare the best refined start with the 8-start reference coupling in objective, angle, and mean shift of a source cell's barycentric target (in tissue radii).}}
\label{tab:qoet-init}
\setlength{\tabcolsep}{3pt}
\resizebox{\columnwidth}{!}{%
\begin{tabular}{llrrrrrrr}
\toprule
Dataset & Interval & Width ($^\circ$) & Other (\%) & Other ($^\circ$) & Replicas & $\Delta J/J$ (\%) & $\Delta\theta$ ($^\circ$) & Shift \\
\midrule
\multirow{4}{*}{Axolotl} & 2$\to$5 & 0.0 & 24.77 & 180 & 9/9 & +0.037 & 1.07 & 0.024 \\
 & 5$\to$10 & 0.0 & 23.07 & 180 & 9/9 & +0.001 & 0.19 & 0.009 \\
 & 10$\to$15 & 0.0 & 13.88 & 180 & 9/9 & -0.005 & 0.78 & 0.022 \\
 & 15$\to$20 & 0.0 & 16.55 & 180 & 9/9 & -0.184 & 1.96 & 0.039 \\
\midrule
\multirow{3}{*}{Chicken} & 4$\to$7 & 109.3 & 0.01 & 20 & 9/9 & +0.000 & 4.06 & 0.058 \\
 & 7$\to$10 & 10.1 & 0.45 & 22 & 9/9 & -0.004 & 1.93 & 0.040 \\
 & 10$\to$14 & 10.6 & 0.42 & 22 & 9/9 & -0.001 & 0.66 & 0.014 \\
\midrule
\multirow{4}{*}{Drosophila} & 5.6$\to$6.3 & 0.7 & 66.14 & 180 & 9/9 & +0.047 & 0.68 & 0.015 \\
 & 6.3$\to$7 & 0.4 & 28.14 & 180 & 9/9 & -0.001 & 0.12 & 0.003 \\
 & 7$\to$9.1 & 0.5 & 11.34 & 178 & 9/9 & -0.001 & 0.19 & 0.005 \\
 & 9.1$\to$10.5 & 0.2 & 22.10 & 180 & 9/9 & -0.019 & 0.22 & 0.006 \\
\midrule
\multirow{4}{*}{Zebrafish} & 5.25$\to$10 & 1.4 & 4.92 & 39 & 9/9 & +0.002 & 0.03 & 0.002 \\
 & 10$\to$12 & 4.4 & 1.65 & 23 & 9/9 & -0.041 & 2.11 & 0.037 \\
 & 12$\to$18 & 3.7 & 4.07 & 23 & 9/9 & -0.064 & 3.39 & 0.047 \\
 & 18$\to$24 & 0.0 & 3.01 & 25 & 9/9 & -0.012 & 1.06 & 0.020 \\
\bottomrule
\end{tabular}}
\end{table}

\rev{On axolotl, Drosophila and zebrafish the pose is well identified.
After refinement, the best solution more than $20^\circ$ away is 1.6--66\% worse, and the starts
within 0.1\% of the best objective span at most $4.4^\circ$.  The best refined
solution agrees with the 8-start coupling to within 0.2\% in objective and
$3.4^\circ$ in angle, and it moves the barycentric target of a source cell by at
most 0.047 tissue radii.  This includes the 5.25--10\,hpf zebrafish interval,
where the embryo is close to rotationally symmetric.  There solutions more than $20^\circ$ away are at least 4.9\% worse and the refined pose differs from the
8-start pose by $0.03^\circ$.  In all nine rotated replicas of the 8-start schedule,
on every interval, the selected start lies in the orientation of the 120-start optimum.

Chicken behaves differently.  Its D4 section has only 147 cells and is nearly
round, and between D4 and D7 the tissue grows 3.6-fold and changes shape.  Over
all orientations the objective then varies by at most 0.2\%, so the two
snapshots alone barely determine the pose.  The next two chicken intervals are better determined (0.4--0.5\% gap, $10$--$11^\circ$ near-optimal width).  To measure what
this means downstream, we rebuilt the D4$\to$D7 coupling from two near-optimal
alternatives, $6^\circ$ and $78^\circ$ from the reference pose (objective
$+0.001\%$ and $+0.10\%$), and retrained with seed 17.  A control that re-solves
the reference pose with the same pipeline gives spatial SW 0.207, Chamfer 0.401
and gene SW 0.580.  The alternatives leave spatial errors unchanged (spatial SW
0.203--0.206, Chamfer 0.396--0.397) but raise gene SW to 0.68--0.69, presumably
because the 147 source cells split among 528 targets and small pose changes reassign
these targets.  This is still below WFR-FM (0.75) and stVCR (1.52) in
Table~\ref{tab:real-data}, and the lowest objective, which the pose search
selects, also gives the lowest molecular error.}

\section{\rev{Cost of the quotient}}
\label{app:runtime-breakdown}

\rev{Relative to Equation~\eqref{eq:wfr-action},
Equation~\eqref{eq:qaction} changes three parts of the implementation
(Table~\ref{tab:runtime-breakdown}).  The coupling becomes a joint search over pose and coupling.  It alternates a weighted Procrustes step, which costs one
pass over the current coupling, with the sparse column-generation solver used at
a fixed pose.  The whole joint search takes 6--15 minutes per series.  The
identity-pose coupling of the no-quotient ablation took longer on chicken and
Drosophila and did not converge within tolerance on axolotl and zebrafish
(Appendix~\ref{app:baseline-details}), but it used the dense
solver, so this comparison does not isolate the cost of the pose
updates.  In training, the rigid fit of
Equation~\eqref{eq:qfm-loss} needs the batch moments and a $2\times2$ solve and
adds at most 11\% to the cost of an update.  In generation, the
projector is recomputed at every RK4 stage from the same moments.  Its cost
is nearly constant in the number of cells at these sizes and adds at most
0.3\,s, or 17\%, to full-trajectory integration.  The projector needs only the first and second
mass-weighted moments of positions and velocities, and the pose update is a
weighted Procrustes problem with a closed-form solution.}

\begin{table}[t]
\centering\small
\caption{\rev{\textbf{Cost of the quotient.}  Training steps and trajectory integrations are medians of alternating repeats on a single dedicated GPU, and the coupling is the wall-clock time of the qOET solves.  \emph{Identity-pose OET} is the coupling of the no-quotient ablation (Table~\ref{tab:quotient-off}), solved densely with the pose fixed.  It did not converge within tolerance on axolotl and zebrafish.  Training steps use the batch sizes of Table~\ref{tab:hparams}, and integrations propagate all cells of the first snapshot.  One call of the projector takes 0.8\,ms for $10^3$--$6.6\times10^4$ cells.}}
\label{tab:runtime-breakdown}
\setlength{\tabcolsep}{3pt}
\resizebox{\columnwidth}{!}{%
\begin{tabular}{lrrrrrr}
\toprule
 & \multicolumn{2}{c}{Coupling (s)} & \multicolumn{2}{c}{Training step (ms)} & \multicolumn{2}{c}{Generation (s)} \\
\cmidrule(lr){2-3}\cmidrule(lr){4-5}\cmidrule(lr){6-7}
Dataset & qOET & identity pose (dense) & horizontal & ambient & projected & ambient \\
\midrule
Axolotl & 794 & not converged & 28.0 & 25.2 & 69.2 & 68.9 \\
Chicken & 466 & 681 & 21.6 & 20.4 & 0.99 & 0.85 \\
Drosophila & 361 & 1,452 & 23.9 & 22.8 & 0.91 & 0.84 \\
Zebrafish & 929 & not converged & 23.9 & 22.4 & 1.87 & 1.64 \\
\bottomrule
\end{tabular}}
\end{table}

\clearpage
\section{\rev{Notation and terms}}
\label{app:notation}

\begin{table}[!ht]
\centering\small
\caption{\rev{\textbf{Notation and terms used in the main text.}}}
\label{tab:notation}
\rev{\begin{tabular}{p{0.24\linewidth}p{0.68\linewidth}}
\toprule
Symbol or term & Meaning \\
\midrule
$z=(x,q)$ & State of a cell, consisting of spatial position $x$ and molecular representation $q$ (32 PCs). \\
$\mu,\nu,\rho_t$ & Tissues as weighted sets of cells, where the weight of a cell is the abundance it represents. \\
$u_x,u_q,g$ & Spatial velocity, molecular velocity and relative growth rate. \\
$\delta,\lambda_x,\lambda_q$ & Cost of mass change, and weights of spatial and molecular displacement. \\
$\WFR_\delta$ & Wasserstein--Fisher--Rao distance, the least action to move cells and change their mass (Eq.~\ref{eq:wfr-action}). \\
$G=\SE(d_x)$, $h=(R,b)$ & Proper rotations $R$ and translations $b$, applied to all positions at once. \\
Orbit $[\mu]$ & All copies of a tissue that differ only by one shared rotation and translation. \\
Rigid field $a+\Omega x$ & Velocity of a moving rigid frame with translation velocity $a$ and rotation rate $\Omega$. \\
Projector $P^H_\rho$ & Removes from a velocity field its best-fitting rigid field, weighted by abundance (Eq.~\ref{eq:projector}). \\
Horizontal & A velocity with no rigid part, so it changes the tissue without moving it as a whole. \\
$\qWFR_\delta$ & Distance between orbits, defined as $\WFR_\delta$ after the best relative pose (Eq.~\ref{eq:qwfr}). \\
qOET, $(h^\star,\gamma^\star)$ & Joint estimate of the relative pose $h^\star$ and the cell matching $\gamma^\star$ (Eq.~\ref{eq:qoet}). \\
Bridge, carrier & Closed-form path of one matched pair, and the weighted point that moves along it (Eqs.~\ref{eq:mass-path}--\ref{eq:teacher}). \\
Quotient / direct metric & Error after / without an evaluation-only rigid alignment of prediction and target. \\
\bottomrule
\end{tabular}}
\end{table}

\end{document}